\documentclass[preprint,12pt, 3p]{elsarticle}

\usepackage{amssymb}
\usepackage{amsmath,amssymb}
\usepackage{bm}
\usepackage{comment}
\usepackage{algorithm, algpseudocode}
\usepackage[subrefformat=parens]{subcaption}
\usepackage{tabularx}
\usepackage{url}

\algnewcommand{\Initialize}[1]{%
  \State \textbf{initialize}
  \Statex \hspace*{\algorithmicindent}\parbox[t]{.8\linewidth}{\raggedright #1}
}

\newcommand\figref[1]{Fig.~\ref{fig:#1}}
\newcommand\tabref[1]{Table~\ref{tab:#1}}
\newcommand\algoref[1]{Algorithm~\ref{algo:#1}}
\renewcommand\eqref[1]{Eq.~(\ref{eq:#1})}
\newcommand\opref[1]{Line~\ref{op:#1}}

\newcommand\figlabel[1]{\label{fig:#1}}
\newcommand\tablabel[1]{\label{tab:#1}}
\newcommand\algolabel[1]{\label{algo:#1}}
\newcommand\eqlabel[1]{\label{eq:#1}}
\newcommand\oplabel[1]{\label{op:#1}}

\newcommand{\argmin}{\mathop{\rm arg~min}\limits} %arg min

\journal{Robotics and Autonomous Systems}

\begin{document}

\begin{frontmatter}



\title{Mirror Descent Search and its Acceleration\tnoteref{t1}}
\tnotetext[t1]{The research was partially supported by JSPS KAKENHI (Grant numbers JP26120005, JP16H03219, and JP17K12737).}


\author[1]{Megumi Miyashita}  
\author[2]{Shiro Yano}
\author[2]{Toshiyuki Kondo}

\address[1]{Dept. of Computer and Information Sciences, Graduate School of Engineering, \\Tokyo University of Agriculture and Technology, Tokyo, Japan}
\address[2]{Division of Advanced Information Technology and Computer Science, Institute of Engineering, \\Tokyo University of Agriculture and Technology, Tokyo, Japan}

\begin{abstract}
In recent years, attention has been focused on the relationship between black-box optimization problem and reinforcement learning problem.
In this research, we propose the Mirror Descent Search (MDS) algorithm which is applicable both for black box optimization problems and reinforcement learning problems. 
Our method is based on the mirror descent method, which is a general optimization algorithm.
The contribution of this research is roughly twofold.
We propose two essential algorithms, called MDS and Accelerated Mirror Descent Search (AMDS), and two more approximate algorithms: Gaussian Mirror Descent Search (G-MDS) and Gaussian Accelerated Mirror Descent Search (G-AMDS).
This research shows that the advanced methods developed in the context of the mirror descent research can be applied to reinforcement learning problem. 
We also clarify the relationship between an existing reinforcement learning algorithm and our method.
With two evaluation experiments, we show our proposed algorithms converge faster than some state-of-the-art methods.
\end{abstract}


\end{frontmatter}


\section{Introduction}
Similarity between black-box optimization problem and reinforcement learning (RL) problem inspires recent researchers to develop novel RL algorithms \cite{Stulp2013Policy, Hwangbo2014ROCKEfficient, Salimans2017Evolution}.
The objective of a black box optimization problem is to find the optimal input $x^*\in\mathcal{X}$ of an unknown function $f: \mathcal{X}\rightarrow\mathbb{R}$.
Because the objective function $f$ is unknown, we usually solve the black box optimization problem without gradient information $\nabla_x f$. 
Such is the case with RL problem. 
The objective of an RL problem is to find the optimal policy that maximizes the expected cumulative reward~\cite{Sutton1998Reinforcement}. 
As is the case in a black-box optimization problem, the agent doesn't know the problem formulation initially, so
he is required to tackle the lack of information. 
%Usually it is assumed that he doesn't have a model of the objective function or a model of plant dynamics.
In this research, we propose RL algorithms from a standpoint of a black-box optimization problem. 

%\textcolor{red}{Model-free approach is one of the well-known approach }
%\textcolor{red}{cannot use optimization techniques. In this research, we propose the algorithms for both of these problems on the basis of mathematical optimization techniques.}
RL algorithm has been categorized into a value-based method and a policy-based method, roughly.
In the value-based method, the agent learns the value function of some action in some state.
On the other hand, in the policy-based method, the agent learns policy from the observation directly.
Moreover, RL algorithm has been divided into a model-free approach and a model-based approach.
In the model-based approach, first, the agent gains the model of a system from the sample. Then, it learns policy or the value using the model.
In contrast, in the model-free approach, the agent learns the policy or value without the model.
RL algorithms usually employ the assumption that the behavior of environment is well approximated by Markov Decision Process (MDP). 

Recently, KL divergence regularization plays a key role in policy search algorithms. 
KL divergence is one of the essential metrics between two distributions. 
Past methods \cite{Schulman2015Trust, Peters2010Relative, Abdolmaleki2017Deriving, Abdolmaleki2015Model, Zimin2013Online, Daniel2012Hierarchical} employ KL divergence regularization to find a suitable distance between a new distribution and a referential distribution. 
It is important to note that there exists two types of KL divergence: KL and reverse-KL (RKL) divergence~\cite{Bishop2006Pattern, nowozin2016f}. 
The past researches mentioned above are clearly divided into 
the algorithms with KL divergence~\cite{Schulman2015Trust, Abdolmaleki2017Deriving} and RKL divergence~\cite{Peters2010Relative, Abdolmaleki2015Model, Zimin2013Online, Daniel2012Hierarchical}. 
We review details of these algorithms afterward. 
% The past researches using KL emphasized the relation with the natural gradient method~\cite{Schulman2015Trust, Abdolmaleki2017Deriving}. 
% On the other hands, as we discuss the relationship between our method that used reverse-KL and the mirror descent.
% We consider that the methods which used reverse-KL~\cite{Peters2010Relative, Abdolmaleki2015Model, Zimin2013Online, Daniel2012Hierarchical} can discuss it same as our method.
% Moreover, they treat multi-modal distribution without performance detriment~\cite{Bishop2006Pattern}.

Bregman divergence is the general metric which includes both of KL and RKL divergence~\cite{Amari2009alpha} (see \ref{sec:alphadiv}). 
Moreover, it includes Euclidean distance, Mahalanobis distance, Hellinger distance and so on. 
Mirror Descent (MD) algorithm employs the Bregman divergence 
to regularize the learning steps of decision variables; it includes a variety of gradient methods~\cite{Bubeck2015Convex}. 
Accelerated mirror descent~\cite{Krichene2015Accelerated} is one of the recent advance applicable for the MD algorithms universally.

In this study, we propose four reinforcement learning algorithms on the basis of MD method. 
Proposed algorithms can be applied in the non-MDP setting. 
%MD is a solution of the optimization problem using the gradient of the objective function.
% We propose two essential algorithms, called mirror descent search(MDS) and accelerated mirror descent search(AMDS), and two more approximate algorithms: Gaussian mirror descent search(G-MDS) and Gaussian accelerated mirror descent search(G-AMDS).
We propose two essential algorithms and two approximate algorithms of them.
We propose mirror descent search (MDS) and accelerated mirror descent search (AMDS) as the essential algorithms, and Gaussian mirror descent search (G-MDS) and Gaussian accelerated mirror descent search (G-AMDS) as the approximate algorithms. 
G-AMDS showed significant improvement in convergence speed and optimality in two benchmark problems. 
If other existing reinforcement learning algorithms can be reformulated as the MDS form, they would also get the benefit from the acceleration. 
We also clarify the relationship between existing reinforcement learning algorithms and our method.
As an example, we show the relationship between MDS and Policy Improvement with Path Integrals (PI${}^\text{2}$) \cite{Theodorou2010generalized, Theodorou2010Reinforcement} in section \ref{sec:MDSPI}.

\section{Related Works}
% \textcolor{red}{1. 本文中ではKL関数を使うが，それはRKLを想定している．これを読者に勘違いさせないように，随所にRKLを使う旨説明する．今回はRKLを使うが，それはAMDの適用によって高速化することを主眼としている．2.NESの段落は，何で微分しているかを伝えたい．我々はノンパラができる．ノンパラベイズの利点を確認．3.Related worksの書き方として，「RKLを使った場合の」MDSはREPSと似ている，という具合に調整する必要あり．→一方で「KLを使った場合の」MDSはTRPOなどに似ているわけだから，あえてREPSだけ取り上げて述べる必要はない．KLもRKLも含んだ一般的な定式化になっている旨述べればよい．4.NESはKLを使っているグループに属するが，重要なのは確率分布をパラメトライズしている点で，我々の手法は原理的にはノンパラでも適用できる．}

This section will proceed in the order described below. 
First of all, we introduce the concept of KL and RKL divergences. 
Then we refer the two types of RL algorithms: RL with KL divergence~\cite{Schulman2015Trust, Abdolmaleki2017Deriving} and RL with RKL divergence~\cite{Peters2010Relative, Abdolmaleki2015Model, Zimin2013Online, Daniel2012Hierarchical}. 
We also refer the RL algorithm PI${}^\text{2}$; we show the relation between PI${}^\text{2}$ and our method afterward. 
We conclude this section with a comment on other MD-based RL algorithms.  

The KL divergence between $\bm x$ and $\bm x'$ is represented as follows. 
\begin{eqnarray}
\mathrm{KL}\left(\bm x, \bm x'\right)=\sum^m_{j=1}x_j\log\frac{x_j}{x'_j} \left(\bm x, \bm x'\in\mathbb{R}^m, x_j, x'_j>0\right).
\end{eqnarray}
% Let $\bm x$ and $\bm x'$ be old distribution and new distribution. 
We call $\mathrm{KL}\left(\bm x', \bm x\right)$ Kullback Leibler divergence under the condition that we determine $\bm x$ by reference to the fixed $\bm x'$; we call $\mathrm{KL}\left(\bm x, \bm x'\right)$ reverse-KL divergence~\cite{nowozin2016f}. 
Bregman divergence includes both of KL and RKL divergence~\cite{Amari2009alpha}, 
so we expect it provides an unified formulation of above-mentioned algorithms.

Let us introduce the RKL-based RL algorithms. 
Relative Entropy Policy Search (REPS) \cite{Peters2010Relative} is one of the pioneering algorithms focusing on the information loss during the policy search process.
The information loss is defined as the relative entropy, also known as the RKL divergence, between the old policy and the new policy. 
The new policy is determined under the upper bound constraints of the RKL divergence. 
Episode-based REPS also considers information loss bound with regard to the upper-level policy~\cite{Daniel2012Hierarchical}.
The method is proposed as an extension of REPS to be an episode-based algorithm.
The paper \cite{Zimin2013Online} discussed the similarity between Episode-based REPS and the proximal point algorithm; they proposed the Online-REPS algorithm as an theoretically guaranteed one. 
MOdel-based Relative Entropy stochastic search (MORE) also employed RKL divergence~\cite{Abdolmaleki2015Model}, which extends the episode-based REPS to be a model-based RL algorithm. 
These algorithms employ RKL divergence in their formulation. 

% There are some methods that are discussed the relationship with REPS.
There are some methods employing KL divergence. 
Trust Region Policy Optimization (TRPO) \cite{Schulman2015Trust}, which is one of the suitable algorithms to solve deep reinforcement learning problem, updates the policy parameters under the KL divergence bound.
The research~\cite{Abdolmaleki2017Deriving} showed that KL divergence between policies plays a key role to derive the well-known heuristic algorithm: Co-variance Matrix Adaptation Evolutionary Strategy (CMA-ES) \cite{Hansen2001Completely}.
Authors named the method Trust-Region Co-variance Matrix Adaptation Evolution Strategy (TR-CMA-ES). 
TR-CMA-ES is similar to episode-based REPS but uses the KL divergence.
Proximal Policy optimization (PPO) algorithm also introduces KL divergence in their penalized objective~\cite{schulman2017proximal}.
% Thus, that causes the research~\cite{Abdolmaleki2017Deriving} to discuss the relationship between CMA-ES and the algorithm used trust region such as REPS.

PI${}^\text{2}$ \cite{Theodorou2010Reinforcement, Theodorou2013information} would be one of the worth mentioning RL algorithm. 
PI${}^\text{2}$ encouraged researchers \cite{Stulp2012Policy,Hwangbo2014ROCKEfficient} to focus on the relationship between RL algorithms and black box optimization. 
For example, \cite{Stulp2012Policy} proposes a reinforcement learning algorithm PI${}^\text{BB}$ on the basis of black box optimization algorithm: CMA-ES.
The authors \cite{Theodorou2012Relative,Theodorou2013information} discussed the connection between PI${}^\text{2}$ and KL control. 
We further discuss PI${}^\text{2}$ from a viewpoint of our proposed methods at section \ref{sec:MDSPI}.

Previous studies also proposed reinforcement learning algorithms on the basis of MD method\cite{Mahadevan2012Sparse, Montgomery2016Guided}. 
Mirror Descent TD($\lambda$) (MDTD) \cite{Mahadevan2012Sparse} is a value based RL algorithm. 
The paper \cite{Mahadevan2012Sparse} employs Minkowski distance with Euclidean space rather than KL divergence. 
By contrast, we basically employ the Bregman divergences on the simplex space, i.e. non-Euclidean space. 
Mirror Descent Guided Policy Search (MDGPS)~\cite{Montgomery2016Guided} is also associated with our proposed method. 
They showed mirror descent formulation improved the Guided Policy Search (GPS) \cite{Levine2013Guided}. 
MDGPS has a distinctive feature that it depends both on KL divergence and RKL divergence. 
However, as is shown in \cite{Krichene2015Efficient}, there are the variety of Bregman divergences on simplex space other than KL divergence and RKL divergence. 
Moreover, it plays an important role in accelerating the mirror descent~\cite{Krichene2015Accelerated}. 
So we explicitly use Bregman divergence in this research. 

\section{Mirror Descent Search and Its Variants}
\subsection{Problem Statement}
In this section, we mainly explain our algorithm as a method for the black box optimization problem.
% この節では，より一般的な表現のために，基本的にはブラックボックス最適化問題との解法としてアルゴリズムを説明する．
Consider the problem of minimizing the original objective function $J$ defined on subspace $\Omega\subseteq\mathbb{R}^l$, i.e. $J: \Omega\rightarrow\mathbb{R}$. 
We represent the decision variable by $\bm\omega\in\Omega$. 
Rather than dealing with decision variable $\bm\omega\in\Omega$ directly, we consider the continuous probability density function of $\bm\omega$.
Let us introduce the probability space. 
The probability space is defined as $\left(\Omega, \mathcal{F}, P\right)$, where $\mathcal{F}$ is the $\sigma$-field of $\Omega$ and $P$ is a probability measure over $\mathcal{F}$.

In this paper, we introduce the continuous probability density function $p(\bm\omega)$ as the alternative decision variable defined on the probability space. 
We also define the alternative objective function by the expectation of the original objective function $J\left(\bm\omega\right)$:
\begin{eqnarray}
\mathcal{J}=\int_\Omega J\left(\bm\omega\right)p\left(\bm\omega\right)d\bm\omega
\end{eqnarray}
Therefore, we search the following domain:
\begin{eqnarray}
p\left(\bm\omega\right)&\geq&0\\
\int_\Omega p\left(\bm\omega\right)d\bm\omega&=&1
\end{eqnarray}
Let us introduce the set  $\mathcal{P}_\text{all}$ consists of all probability density functions defined on the probability space.  The optimal generative probability is
\begin{equation}
p^*(\bm\omega)=\argmin_{p(\bm\omega)\in\mathcal{P}_\text{all}}\left\{\int_\Omega J\left(\bm\omega\right)p\left(\bm\omega\right)d\bm\omega\right\}=\argmin_{p(\bm\omega)\in\mathcal{P}_\text{all}}\mathcal{J}.
\end{equation}
From the viewpoint of the black box optimization problems, the algorithm aims at obtaining the optimal decision variable $p^*(\bm\omega)$ to optimize the alternative objective function $\mathcal{J}$.
From the viewpoint of the reinforcement learning problems, it's purpose is to obtain the optimal policy $p^*(\bm\omega)$ to optimize reward $\mathcal{J}$.
Next, we introduce an iterative algorithm converges to the optimal solution.
\subsection{Mirror Descent Search and Gaussian-Mirror Descent Search}
\subsubsection{Mirror Descent Search (MDS)}
The algorithm is divided into three steps as \figref{update}.
% アルゴリズム全体の流れは\figref{update}のように，3つのステップで表される．本節ではそれぞれについて説明する
\begin{figure}[htb]
\centering
\includegraphics[clip,width=.6\linewidth]{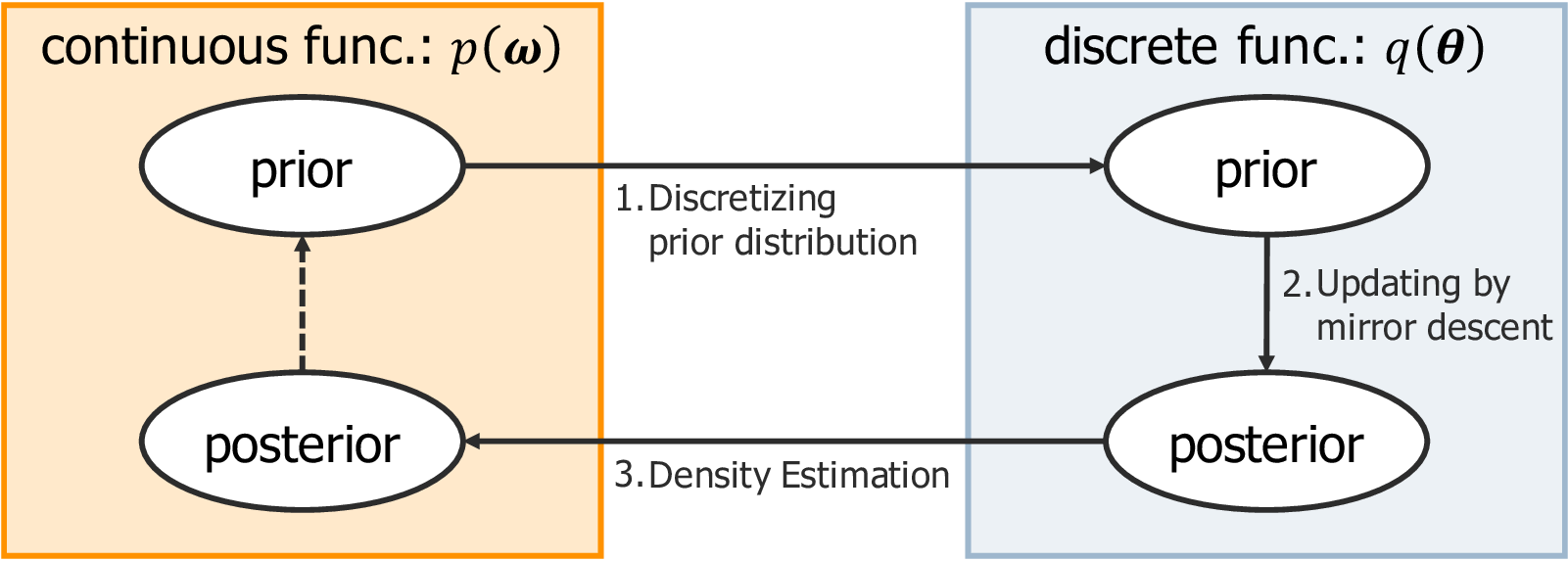}
\caption{Rough scheme of mirror descent search.}\figlabel{update}
\end{figure}
\subsubsection*{Discretizing Prior Distribution}
To update the continuous probability density function $p\left(\bm\omega\right)$, we need to discretize the probability density function from sampling, because we don’t know the form of the objective function. 
We can only evaluate the objective value corresponding to each sample.
% 目的関数の関数の形が不明で積分ができないため，サンプリングによって離散化して考える．

First, we discretize $p\left(\bm\omega\right)$ based on sampling.
% まず，サンプリングに基づいて近似的に$p\left(\bm\omega\right)$を離散化する．
For the illustrative purpose, we assume that we can get infinite samples $\bm\theta_i \sim p\left(\bm\omega\right)$, here.
% $p\left(\bm\omega\right)$から無限個のサンプリングを行うと仮定する．
% Corresponds to the samples $\bm\theta\in\Omega$, the function $q:\Omega\rightarrow\mathbb{R}$ is defined.
To satisfy the definition of the discrete probability density function, we discretize the continuous distribution $p$ using the function $q:\Omega\rightarrow\mathbb{R}$ for the acquired samples~\cite{Billingsley2008Probability}:
% そしてサンプリングされたパラメータ$\left\{\bm\theta_1, \dots\right\}=\Theta\in\Omega$において，関数$q:\Theta\rightarrow\mathbb{R}$を定義する．この$q$は離散のprobability density functionの条件を満たす.
\begin{eqnarray}
\eqlabel{discretizedq}
q\left(\bm\theta_i\right)&:=&\lim_{\Delta\bm\theta\rightarrow0}\frac{p\left(\bm\theta_i\leq\bm\omega\leq\bm\theta_i+\Delta\bm\theta\right)}{\sum_{j=0}^{\infty}p\left(\bm\theta_j\leq\bm\omega\leq\bm\theta_j+\Delta\bm\theta\right)}\ \left(1\leq i\leq\infty\right)\\
\sum_{j=0}^{\infty}q\left(\bm\theta_j\right)&=&1.%, q\left(\bm\theta_i\right)\geq0%\ \left(1\leq i\leq\infty\right)
\eqlabel{discretizedq_condition}
\end{eqnarray}
With \eqref{discretizedq} and \eqref{discretizedq_condition}, our objective function $\tilde{\mathcal{J}}$ becomes the expectation of the original objective function $\bm J$: 
\begin{eqnarray}
\tilde{\mathcal{J}}=\sum^\infty_{j=1}J\left(\bm\theta_j\right)q\left(\bm\theta_j\right)=\langle{\bm J}, \bm q\rangle, 
\end{eqnarray}
where %we represent the decision variable vector with $\bm q$, and the cost vector with $\bm J$.
\begin{eqnarray}
\bm q=\left[q_1, \dots\right]&:=&\left[q\left(\bm\theta_1\right), \dots\right]\in\mathcal{Q}\\
\bm J=\left[J_1, \dots\right]&:=&\left[J\left(\bm\theta_1\right), \dots\right]\in\mathbb{R}^\infty.
\end{eqnarray}

% Thus, our optimization problem becomes
% % Therefore, the optimal generative probability is
% \begin{equation}
% \bm q^*=\argmin_{\bm q\in\mathcal{Q}}\tilde{\mathcal{J}}=\argmin_{\bm q\in\mathcal{Q}}\langle{\bm J}, \bm q\rangle.
% \end{equation}

\subsubsection*{Updating by Mirror Descent}
% Next, we obtain the optimal policy by updating $\bm q$.
% As a means for updating $\bm q$, we use the Mirror Descent(\ref{sec:MirrorDescent}):
After discretizing the continuous distribution $p_{k-1}(\bm \omega)$, we employ the mirror descent algorithm (\ref{sec:MirrorDescent}) to update the discretized distribution $\bm q_{k-1}$: 
\begin{equation}
\eqlabel{RewriteMirrorDescentInfD}
\bm q_k=\argmin_{\bm q\in\mathcal{Q}}\left\{\langle\nabla_{\bm q}\tilde{\mathcal{J}}, \bm q\rangle+\eta B_\phi\left(\bm q, \bm q_{k-1}\right)\right\},
\end{equation}
where $\eta$ is step-size.
We call $\bm q_{k-1}$ as the prior distribution, and $\bm q_{k}$ as the posterior distribution. 
The domain of the decision variable $\bm q$ is the simplex $\mathcal{Q}$.
$B_\phi$ is the Bregman divergence, which has an arbitrarily smooth convex function $\phi$ and is defined as 
\begin{equation}
\eqlabel{BregmanDivergence}
B_\phi\left(x, x'\right)=\phi\left(x\right)-\phi\left(x'\right)-\langle\nabla\phi\left(x'\right), x-x'\rangle.
\end{equation}
% $\bm q_k$ is calculated from prior distribution $\bm q_{k-1}$.
There are numerous variations of Bregman divergence on the simplex such as the KL divergence $\phi\left(x_k\right)=\sum^M_{j=1}x_{k,j}\log\left(x_{k,j}\right)$ and the Euclidean distances assumed on the simplex \cite{Krichene2015Efficient}. Moreover, slightly perturbed KL divergence, which was first introduced in \cite{Krichene2015Efficient}, is another important divergence. 
It plays a key role in accelerating the convergence speed of mirror descent as discussed in \cite{Krichene2015Accelerated} and this paper. 

% Regarding the decision variable $\bm q$ as the parameter in the original MD, we obtain the following:
Because $\nabla_{\bm q}\tilde{\mathcal{J}}=\bm J$, we finally obtain the convex optimization problem:
% We can derive:
% \begin{eqnarray}
% \nabla_{\bm q}\tilde{\mathcal{J}}
% &=&\frac{\partial\tilde{\mathcal{J}}}{\partial\bm q}=\frac{\partial\langle{\bm J}, \bm q\rangle}{\partial\bm q}\nonumber\\
% &=&\left[\frac{\partial J(\bm\theta_1)q(\bm\theta_1)+\partial J(\bm\theta_2)q(\bm\theta_2)+\dots}{\partial q(\bm\theta_1)},
% \frac{\partial J(\bm\theta_1)q(\bm\theta_1)+\partial J(\bm\theta_2)q(\bm\theta_2)+\dots}{\partial q(\bm\theta_2)},
% \cdots\right]\nonumber\\
% &=&\left[ J\left(\bm\theta_1\right),  J\left(\bm\theta_2\right), \cdots\right]=\bm J.
% \eqlabel{nabla}
% \end{eqnarray}
% That is, $\nabla_{\bm q}\tilde{\mathcal{J}}$ is a value obtained without using derivatives of $\bm J$. From the above, $\bm q_k$ can be updated using \eqref{RewriteMirrorDescentInfD} and learning can proceed.
% 純粋なブラックボックス最適化など，実時間の概念がない環境である場合，the alternative decision variableの更新はここまでの説明で完結し， \eqref{RewriteMirrorDescentInfD} is explained as following equation:
\begin{eqnarray}
\bm q_k=\argmin_{\bm q\in\mathcal{Q}}\left\{\langle\bm J, \bm q\rangle+\eta B_\phi\left(\bm q, \bm q_{k-1}\right)\right\}.\eqlabel{MDSJ}
\end{eqnarray}

Although we have assumed the infinite number of samples from $p$, it works only in theory. 
% ここまでは$p$から無限個サンプリングすることを仮定していたが，実装の都合上，実際には無限個サンプリングすることはできない．
In what follows, we approximate the distribution $\bm q$ using sufficiently large $m$ samples.
% そのためここからは十分に大きな数$m$として，$m$個サンプリングすることを考える．
% In order to solve this problem, we must generate trajectories of $n$ kinds for $m$-type policy parameters, and make $m\times n$ attempts for one update. Here, we use the concept of online learning.
% $r\left(\bm\tau^{\bm\theta_i}\right)$の分散が限りなく$0$に近い場合，we can approximate $\sum^n_{j=1}r\left(\left(\bm\tau^{\bm\theta_i}\right)_j\right)$ by $r\left(\left(\bm\tau^{\bm\theta_i}\right)_j\right)$ for any $j$.
% \textcolor{blue}{「分散が0に近い場合は近似誤差が小さくなる」議論が適切か要検討．}
% When the variance of $r\left(\bm\tau^{\bm\theta_i}\right)$ is close to 0 as possible, we can approximate $\sum^n_{j=1}r\left(\left(\bm\tau^{\bm\theta_i}\right)_j\right)$ by $r\left(\left(\bm\tau^{\bm\theta_i}\right)_j\right)$ for any $j$.

% Considering the above, the gradient of the alternative objective function is derived as follows:
% \begin{equation}
% \eqlabel{ObjFuncOnline}
% \nabla_{\bm q}\tilde{\mathcal{J}}\propto \left[j\left(\tau^{\bm\theta_1}\right),\cdots,j\left(\tau^{\bm\theta_m}\right)\right]:=\bm j.
% \end{equation}
% Thus, $\left[j\left(\tau^{\bm\theta_1}\right),\cdots,j\left(\tau^{\bm\theta_m}\right)\right]$ that is a vector of the cumulative reward before calculating the expected value, can be used as the gradient of MD:
% \begin{eqnarray}
% \bm q_k=\argmin_{\bm q\in\mathbb{R}^\infty}\left\{\langle\bm j_{k-1}, \bm q\rangle+\eta_kB_\phi\left(\bm q||\bm q_{k-1}\right)\right\}.
% \end{eqnarray}
% Thus, we evaluate once for each step and rollout.

% Because this derived algorithm is a policy search based on MD, it is called MDS.

\subsubsection*{Density Estimation}
% 更新後のベクトルを確率変数，確率をサンプル数と考えて，連続のprobability density functionを推定する．
We estimate the continuous probability density function $p_k(\bm \omega)$ from the posterior distribution $\bm q_{k}$. 
% , by considering the updated vector as a stochastic variable and probability as the number of samples.
The procedure of MDS with $K$-iterations is summarized in \algoref{MDS}.

\begin{algorithm}
\caption{Mirror descent search}\algolabel{MDS}      
\begin{algorithmic}[1]
\Initialize{continuous functions: $p_0\left(\bm\omega\right):=p_\text{init.}\left(\bm\omega\right).$
}
\For{$k=1$ to $K$}
    \For{$i=1$ to $m$}
        \State Sample parameter $\bm\theta_{i}\sim p_{k-1}(\bm\omega)$.\oplabel{mds:sample}
        \State (Discretize $p_{k-1}$) $q_{k-1,i}=q(\bm\theta_{i})$.\oplabel{mds:calcq}
        \State (Evaluate) $J_{k-1,i}=J(\bm\theta_{i})$.\oplabel{mds:calcJ}
    \EndFor
    \State $\hat{\bm q}_k=\argmin_{\bm q\in\mathbb{R}^m}\left\{\langle\bm J_{k-1}, \bm q\rangle+\eta B_\phi\left(\bm q, \bm q_{k-1}\right)\right\}.$\oplabel{mds:update}
    \State Estimate continuous functions $p_k(\bm\omega)$ from $\hat{\bm q}_k$.\oplabel{mds:estimate}
\EndFor
\end{algorithmic}
\end{algorithm}

\subsubsection{Gaussian-Mirror Descent Search (G-MDS)}
We consider a specific case where the Bregman divergence $B_\phi$ in \eqref{RewriteMirrorDescentInfD} is the RKL divergence. 
Then, \eqref{MDSJ} can be rewritten as follows:
\begin{eqnarray}
\bm q_k=\argmin_{\bm q\in\mathcal{Q}}\left\{\langle\bm J, \bm q\rangle+\eta\mathrm{KL}\left(\bm q, \bm q_{k-1}\right)\right\}.\eqlabel{GMDSJ}
\end{eqnarray}
In G-MDS, we considered $q_{k,i}=q\left(\bm\theta_{k,i}\right)$ as the Gaussian distribution of the mean $\bm\mu_{k-1}\in\mathbb{R}^l$ and the variance-covariance matrix $\bm\Sigma_{\epsilon_{k-1}}\in\mathbb{R}^{l\times l}$, so $\bm\theta_{k, i}$ is generated accordingly:
\begin{equation}
\eqlabel{Gauss1}
% q_k\left(\bm\theta_i\right)=\mathcal{N}\left(\bm\theta\mid\bm\mu_{k-1}, \bm\Sigma_{\epsilon_{k-1}}\right)
\bm\theta_{k, i}\sim\mathcal{N}\left(\bm\mu_{k-1}, \bm\Sigma_{\epsilon_{k-1}}\right)
\end{equation}

Because the derived algorithm is an instance of MDS with the constraint that the policy is a Gaussian distribution, we named G-MDS.
The procedure of G-MDS with $K$-iterations is summarized in \algoref{GMDS}.

As shown in section \ref{sec:MDSPI} and section \ref{sec:eval}, we discuss G-MDS formulation sheds new light on the existing method PI$^\text{2}$. 
Deisenroth also discussed the similarity between episode-based REPS and PI$^\text{2}$\cite{Deisenroth2013Survey}. 
To compare the asymptotic behavior of these algorithms appropriately, \algoref{GMDS} only update the mean vector of Gaussian distribution as PI$^\text{2}$ also only updates the mean vector. 
A lot of past studies proposed the procedure to update variance-covariance matrix \cite{Hansen2001Completely, Stulp2012Path, Abdolmaleki2017Deriving}. 
These methods would be applicable to the G-MDS. 
% \textcolor{blue}{TODO====================}
% originalなPI2やREPSとの関係を評価するために平均値の更新のみ考える．
% 分散は様々なアップデートの仕方がある。今回は定数対角行列として扱う。
% 高次元小標本だから云々という論の立て方は可能？

\begin{algorithm}
\caption{Gaussian mirror descent search}\algolabel{GMDS}      
\begin{algorithmic}[1]
\Initialize{continuous Gaussian function: $p_0\left(\bm\omega\right):=p_\text{init.}\left(\bm\omega\right)$\\
               variance: $\bm\Sigma$.
}
\For{$k=1$ to $K$}
    \For{$i=1$ to $m$}
        \State Sample parameter $\bm\theta_{i} \sim p_{k-1}(\bm\omega)$.\oplabel{gmds:sample}
        \State (Discretize $p_{k-1}$) $q_{k-1,i}=q(\bm\theta_{i})$.\oplabel{gmds:calcq}
        \State (Evaluate) $J_{k-1,i}=J(\bm\theta_{i})$.\oplabel{gmds:calcJ}
    \EndFor
    \State $\hat{\bm q}_k=\argmin_{\bm q\in\mathbb{R}^m}\left\{\langle\bm J_{k-1}, \bm q\rangle+\eta\mathrm{KL}\left(\bm q, \bm q_{k-1}\right)\right\}.$\oplabel{gmds:update}
    \State Estimate the mean $\tilde{\bm\mu}_k$ from $\hat{\bm q}_k$.\oplabel{gmds:estimate}
    \State Generate continuous function $p_k(\bm\omega)$ from $\bm\mu_k$ and $\bm\Sigma$.\oplabel{gmds:generate}
\EndFor
\end{algorithmic}
\end{algorithm}

\subsection{Accelerated Mirror Descent Search and Gaussian-Accelerated Mirror Descent Search}
\subsubsection{Accelerated Mirror Descent Search (AMDS)}
Next, the accelerated mirror descent (AMD) method \cite{Krichene2015Accelerated} is applied to the proposed method. AMD is an accelerated method that generalizes Nesterov's accelerated gradient such that it can be applied to MD.
The details of AMD are explained in \ref{sec:AcceleratedMirrorDescent}.
Here, AMD yields the following equations:
\begin{eqnarray}
\eqlabel{AMDSx}
\bm q_{k}&=&\lambda_{k-1}\bm q_{k-1}^{\tilde z}+\left(1-\lambda_{k-1}\right)\bm q_{k-1}^{\tilde x}, \text{with } \lambda_{k-1}=\frac{r}{r+(k-1)}\\
\eqlabel{AMDSztilde}
\bm q_{k}^{\tilde z}&=&\argmin_{\bm q^{\tilde z}\in\mathbb{R}^m}\left\{\frac{(k-1)s}{r}\langle \bm J_{k-1}, \bm q^{\tilde z}\rangle+B_\phi\left(\bm q^{\tilde z}, \bm q^{\tilde z}_{k-1}\right)\right\}\\
\eqlabel{AMDSxtilde}
\bm q_k^{\tilde x}&=&\argmin_{\bm q^{\tilde x}\in\mathbb{R}^m}\left\{\gamma s\langle \bm J_{k-1}, \bm q^{\tilde x}\rangle+R\left(\bm q^{\tilde x}, \bm q_{k}\right)\right\}
\end{eqnarray}
where $R$ is regularization function, which belongs to the Bregman divergence~\cite{Krichene2015Accelerated}, $r$ and $\gamma$ are hyper parameters, and $s$ is step-size.

The procedure of AMDS with $K$-iterations is summarized in \algoref{AMDS}.
\figref{AMDasRL} also explains the implementation of AMDS. 
Each captions in \figref{AMDasRL} correspond to the line number of \algoref{AMDS}.

\begin{algorithm}
\caption{Accelerated mirror descent search}\algolabel{AMDS}      
\begin{algorithmic}[1]
\Initialize{continuous functions: $p_0^{\tilde z}\left(\bm\omega\right):=p_\text{init.}^{\tilde z}\left(\bm\omega\right), p_0^{\tilde x}\left(\bm\omega\right):=p_\text{init.}^{\tilde x}\left(\bm\omega\right).$
%               vectors: $\bm q_0^{\tilde z}, \bm q_0^{\tilde x}$ by sampling from the continuous functions.\\
}
\For{$k=1$ to $K$}
    \State $p_{k}(\bm\omega)=\lambda_{k-1}p_{k-1}^{\tilde z}(\bm\omega)+\left(1-\lambda_{k-1}\right)p_{k-1}^{\tilde x}(\bm\omega), \text{with } \lambda_{k-1}=\frac{r}{r+(k-1)}$.\oplabel{amds:x}
    \For{$i=1$ to $m$}
        \State Sample parameter $\bm\theta_{i} \sim p_{k}(\bm\omega)$.\oplabel{amds:sample}
        \State (Discretize $p^{\tilde z}_{k-1}$) $q^{\tilde z}_{k-1,i}=p_{k-1}^{\tilde z}(\bm\theta_{i})$.\oplabel{amds:calcqz}
        \State (Discretize $p_{k-1}^{\tilde x}$) $q_{k-1,i}^{\tilde x}=p_{k-1}^{\tilde x}(\bm\theta_{i})$.\oplabel{amds:calcqx}
        \State (Evaluate) $J_{k-1,i}=J(\bm\theta_{i})$.\oplabel{amds:calcJ}
    \EndFor
    \State ${\bm q_{k}}=\lambda_{k-1}{\bm q^{\tilde z}_{k-1}+(1-\lambda_{k-1}){\bm q_{k-1}^{\tilde x}} }$
    \State $\hat{\bm q}_k^{\tilde z}=\argmin_{\bm q^{\tilde z}\in\mathbb{R}^m}\left\{\frac{(k-1)s}{r}\langle \bm J_{k-1}, \bm q^{\tilde z}\rangle+B_\phi\left(\bm q^{\tilde z}, \bm q^{\tilde z}_{k-1}\right)\right\}$\oplabel{amds:ztilde}
    \State $\hat{\bm q}_k^{\tilde x}=\argmin_{\bm q^{\tilde x}\in\mathbb{R}^m}\left\{\gamma s\langle \bm J_{k-1}, \bm q^{\tilde x}\rangle+R\left(\bm q^{\tilde x}, \bm q_{k}\right)\right\}$\oplabel{amds:xtilde}
    \State Estimate continuous functions $p_k^{\tilde z}(\bm\omega), p_k^{\tilde x}(\bm\omega)$ from $\hat{\bm q}_k^{\tilde z}, \hat{\bm q}_k^{\tilde x}$.\oplabel{amds:estimate}
\EndFor
\end{algorithmic}
\end{algorithm}

\begin{figure}[htb]
\centering
\begin{tabular}{ccc}
  \begin{minipage}[b]{.3\linewidth}
    \centering
    \includegraphics[clip,width=1\linewidth]{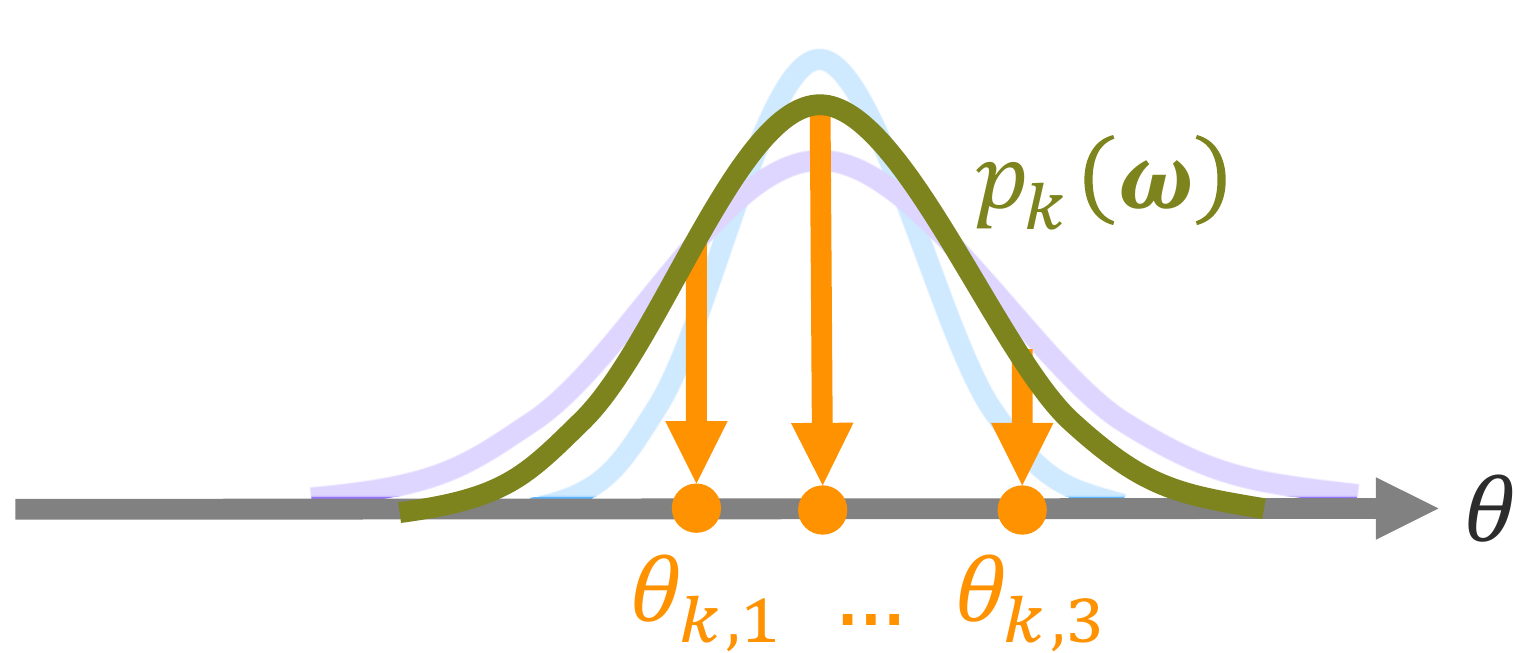}
    \subcaption{\opref{amds:sample}: Get samples\\ $\bm \theta_i\sim p_k(\omega)$}
  \end{minipage}
  \begin{minipage}[b]{.3\linewidth}
    \centering
    \includegraphics[clip,width=1\linewidth]{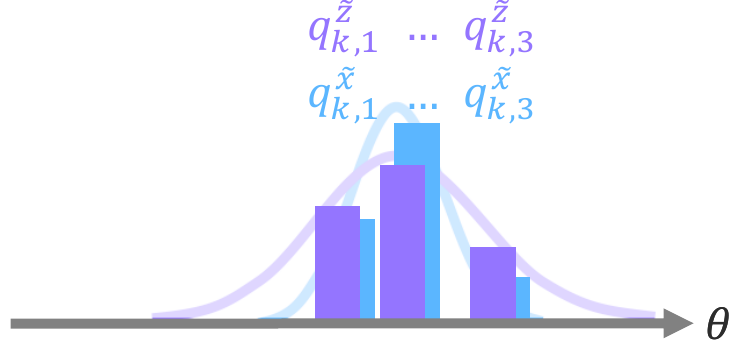}
    \subcaption{\opref{amds:calcqz}: Discretize distributions by substituting $\bm\theta_i$}
  \end{minipage}
  \begin{minipage}[b]{.3\linewidth}
    \centering
    \includegraphics[clip,width=1\linewidth]{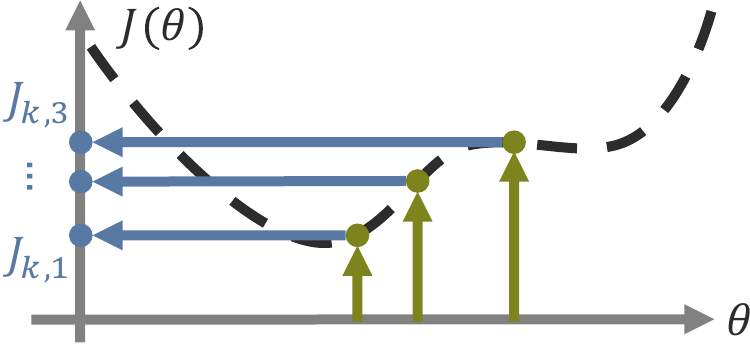}
    \subcaption{\opref{amds:calcJ}: Evaluate $J(\bm\theta_i)$}
  \end{minipage}\\
  \begin{minipage}[b]{.3\linewidth}
    \centering
    \includegraphics[clip,width=1\linewidth]{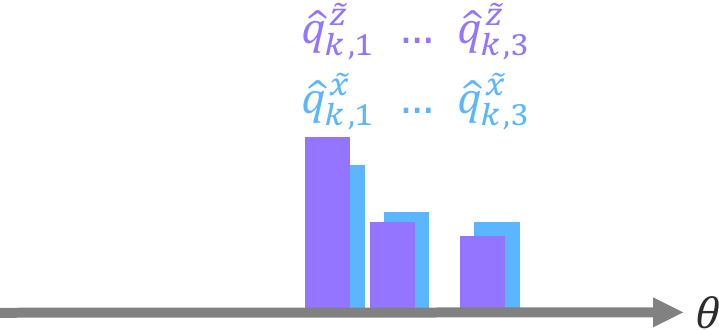}
    \subcaption{Lines~\ref{op:amds:ztilde}--\ref{op:amds:xtilde}: Update the discretized distributions}
  \end{minipage}
  \begin{minipage}[b]{.3\linewidth}
    \centering
    \includegraphics[clip,width=1\linewidth]{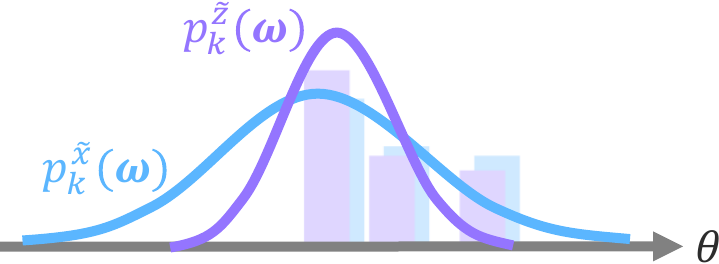}
    \subcaption{\opref{amds:estimate}: Estimate continuous distributions}
  \end{minipage}
  \begin{minipage}[b]{.3\linewidth}
    \centering
    \includegraphics[clip,width=1\linewidth]{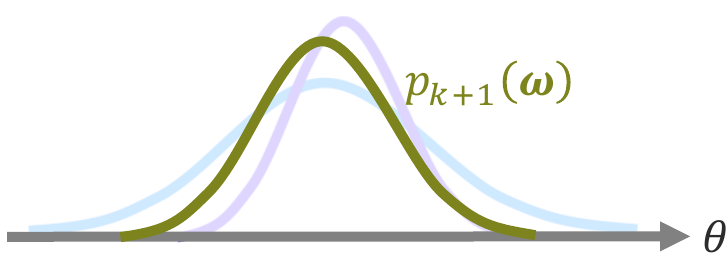}
    \subcaption{\opref{amds:x}: Aggregate two continuous distributions}
  \end{minipage}
  \end{tabular}
  \caption{Visualization of \algoref{AMDS}}\figlabel{AMDasRL}
  % \textcolor{blue}{TODO: 表現を考える}
\end{figure}

\subsubsection{Gaussian-Accelerated Mirror Descent Search (G-AMDS)}
% We derive the same procedure as G-MDS. 

In accordance with prior work \cite{Krichene2015Accelerated}, we applied the RKL distance to the Bregman divergence $ B_\phi $ in \eqref{AMDSztilde} and $\psi\left(x\right)=\varepsilon\sum_{i=1}^n\left(x_i+\varepsilon\right)\log\left(x_i+\varepsilon\right)\left(\bm x\in\mathbb{R}^m, x_{t,j}>0\right)$ on $R=B_\psi$ in \eqref{AMDSxtilde}. 
As the divergence $R$ takes the form of slightly perturbed KL divergence, we represent $R$ by $\mathrm{KL}_\varepsilon$ in \algoref{GAMDS}. 
We approximate the distributions $p^{\tilde x}\left(\bm\theta\right)$ and $p^{\tilde z}\left(\bm\theta\right)$ with a Gaussian distribution.
Accordingly, this method is called G-AMDS. 
Although the result cannot be calculated analytically, it is known that an efficient numerical calculation of $\mathcal{O}(m\log m)$ time is available\cite{Krichene2015Accelerated}.
The procedure of G-AMDS with $K$-iterations is summarized in \algoref{GAMDS}.
\begin{algorithm}
\caption{Gaussian accelerated mirror descent search}\algolabel{GAMDS}      
\begin{algorithmic}[1]
\Initialize{continuous Gaussian functions: $p_0^{\tilde z}\left(\bm\omega\right):=p_\text{init.}^{\tilde z}\left(\bm\omega\right), p_0^{\tilde x}\left(\bm\omega\right):=p_\text{init.}^{\tilde x}\left(\bm\omega\right).$\\
               variance: $\bm\Sigma^{\tilde z}, \bm\Sigma^{\tilde x}.$
%               number of updates: $K$.
}
\For{$k=1$ to $K$}
    \State $p_{k}(\bm\omega)=\lambda_{k-1}p_{k-1}^{\tilde z}(\bm\omega)+\left(1-\lambda_{k-1}\right)p_{k-1}^{\tilde x}(\bm\omega), \text{with } \lambda_{k-1}=\frac{r}{r+(k-1)}$.\oplabel{gamds:x}
    \For{$i=1$ to $m$}
        \State Sample parameter $\bm\theta_{i} \sim p_{k}(\bm\omega)$.\oplabel{gamds:sample}
        \State (Discretize $p^{\tilde z}_{k-1}$) $q^{\tilde z}_{k-1,i}=p^{\tilde z}_{k-1}(\bm\theta_{i})$.\oplabel{gamds:calcqz}
        \State (Discretize $p_{k-1}^{\tilde x}$) $q_{k-1,i}^{\tilde x}=p_{k-1}^{\tilde x}(\bm\theta_{i})$.\oplabel{gamds:calcqx}
        \State (Evaluate) $J_{k-1,i}=J(\bm\theta_{i})$.\oplabel{gamds:calcJ}
    \EndFor
    \State ${\bm q_{k}}=\lambda_{k-1}{\bm q^{\tilde z}_{k-1}+(1-\lambda_{k-1}){\bm q_{k-1}^{\tilde x}} }$
    \State $\hat{\bm q}_k^{\tilde z}=\argmin_{\bm q^{\tilde z}\in\mathbb{R}^m}\left\{\frac{(k-1)s}{r}\langle \bm J_{k-1}, \bm q^{\tilde z}\rangle+\text{KL}\left(\bm q^{\tilde z}, \bm q^{\tilde z}_{k-1}\right)\right\}$\oplabel{gamds:ztilde}
    \State $\hat{\bm q}_k^{\tilde x}=\argmin_{\bm q^{\tilde x}\in\mathbb{R}^m}\left\{\gamma s\langle \bm J_{k-1}, \bm q^{\tilde x}\rangle+\mathrm{KL}_\varepsilon\left(\bm q^{\tilde x}, \bm q_{k}\right)\right\}$\oplabel{gamds:xtilde}
    \State Estimate the means $\bm\mu^{\tilde z}_k, \bm\mu^{\tilde x}_k$ from $\hat{\bm q}_k^{\tilde z}, \hat{\bm q}_k^{\tilde x}$.\oplabel{gamds:estimate}
    \State Generate continuous functions $p_k^{\tilde z}(\bm\omega), p_k^{\tilde x}(\bm\omega)$ from $\bm\mu^{\tilde z}_k, \bm\mu^{\tilde x}_k, \bm\Sigma^{\tilde z}, \bm\Sigma^{\tilde x}$.\oplabel{gamds:generate}
\EndFor
\end{algorithmic}
\end{algorithm}

\section{Experimental Evaluations}
\label{sec:eval}
In this section, we show the comparative experiments.
% 本節では比較実験とその結果について述べる．
% なぜPI2とepisode-based REPS を比較対象にしたか考える：提案手法に近く，シンプルな手法かつstate-of-the-artだから．
We compare the learning curves of G-MDS, G-AMDS, PI${}^\text{2}$ and episode-based REPS in two tasks.
We selected PI${}^\text{2}$ and episode-based REPS as the baseline because they are state-of-the-art methods.
% 2種類のタスクにおいて提案手法G-MDS, G-AMDSと先行手法PI${}^\text{2}$の比較を行う．
In \cite{Theodorou2010generalized, Theodorou2010Reinforcement}, these methods equipped the heuristics such as the normalization of the costs and the simulated annealing.
However, in our evaluations, we do not use these heuristics. We focus on the theoretical guaranteed performance of these algorithms.
% \cite{Theodorou2010generalized, Theodorou2010Reinforcement} ではコストの正規化・ノイズの分散のアニーリングを行なっているが，性能評価においてヒューリスティックを排除するために本節の実験では行っていない．
Our source code is available online\footnote{\url{https://github.com/mmilk1231/MirrorDescentSearch}\ \ We acknowledge with appreciation that PI${}^\text{2}$ code\footnote{} \cite{Theodorou2010generalized} and the AMD code\footnote{} \cite{Krichene2015Accelerated} are gratefully helpful to implement our code.}.
\footnotetext[2]{\url{http://www-clmc.usc.edu/software/git/gitweb.cgi?p=matlab/pi2.git}}
\footnotetext[3]{\url{https://github.com/walidk/AcceleratedMirrorDescent}}

\subsection{2DOF Point Via-point Task}
We performed a 2DOF point via-point task to evaluate the proposed method. The agent is represented as a point on the x--y plane. 
This agent learns to pass through the point (0.5, 0.2) at 250 ms. 
We employed DMP \cite{Ijspeert2003Learning} to parameterize the policy. 
DMP represents the trajectory of agent behavior toward x-axis and y-axis in each time step. 
The parameter settings are as follows: 100 updates, 10 rollouts, and 20 basis functions.
Before learning, an initial trajectory from (0, 0) to (1, 1) is generated. 

The reward function is as follows:
\begin{eqnarray}
r_t &=& 5000f_t^2+0.5\bm\theta^{\rm T}\bm\theta\\
\Delta r_{\rm 250ms}&=&1.0\times 10^{10}\left(\left(0.5-x_{\rm 250ms}\right)^2+\left(0.2-y_{\rm 250ms}\right)^2\right),
\end{eqnarray}
where $\bm \theta\in\mathbb{R}^{20}$ denotes the policy parameter. 

We summarize the results in \figref{Point}. 
\figref{PointCost} shows that G-AMDS agent learns faster than all the other agents. 
% タスクも達成できていることがわかる．
\figref{PointTraj} shows that the agent was able to accomplish the task.

\tabref{PointConv} shows the average cost and the standard deviation of the cost at the last update (right-endpoint of \figref{PointCost}). 
In the figure, the thin line represents a standard deviation of the cost ($\pm\sigma$).
\figref{PointTraj} shows the acquired trajectory at the last update.
We set the variance-covariance matrix of sampling distribution to the unit matrix in all algorithms.

\begin{figure}[htb]
\centering
  \begin{minipage}[b]{.49\linewidth}
    \centering
    \includegraphics[clip,width=1\linewidth]{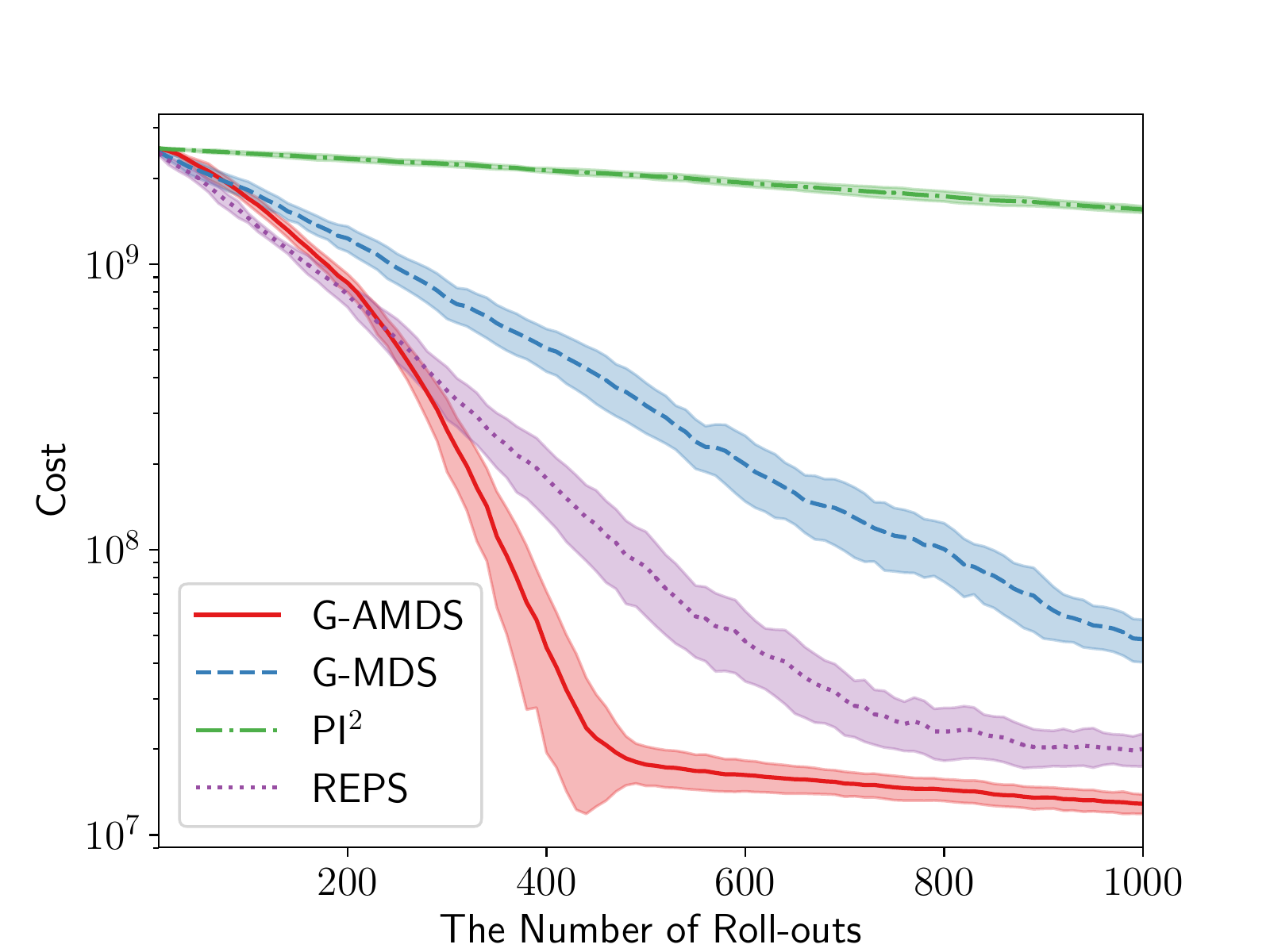}
    \subcaption{Cost}\figlabel{PointCost}
  \end{minipage}
  \begin{minipage}[b]{.49\linewidth}
    \centering
    \includegraphics[clip,width=1\linewidth]{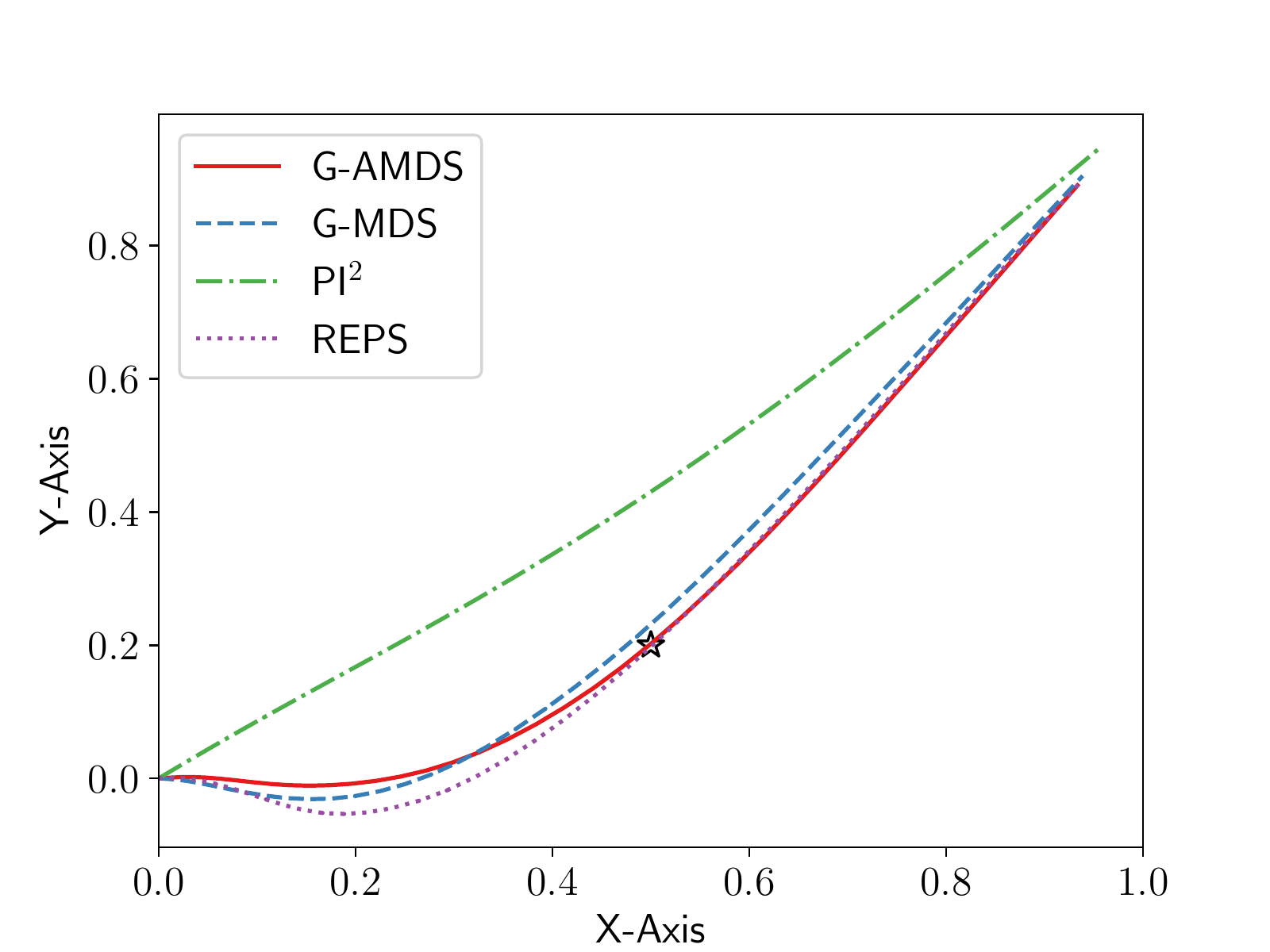}
    \subcaption{Trajectory}\figlabel{PointTraj}
  \end{minipage}
  \caption{2DOF point via-point task}\figlabel{Point}
\end{figure}

\begingroup
\renewcommand{\arraystretch}{2}
\newcolumntype{C}{>{\centering\arraybackslash}X}
\begin{table}[htb]
  \caption{Final cost of 2DOF point via-point task}\tablabel{PointConv}
  \begin{tabularx}{\linewidth}{l|C}
    G-AMDS & $1.3\times10^7 \pm 1.0\times10^6$\\
    G-MDS & $4.9\times10^7 \pm 8.3\times10^6$\\
    PI${}^\text{2}$ & $1.6\times10^9 \pm 4.1\times10^7$\\
    Episode-based REPS  & $2.0\times10^7 \pm 2.6\times10^6$\\
  \end{tabularx}
\end{table}
\endgroup

\subsection{10DOF Arm Via-point Task and 50DOF Arm Via-point Task}
We performed a 10DOF arm via-point task and a 50DOF arm via-point task to evaluate the proposed method. 
The agent learns to control his end-effector to pass through the point (0.5, 0.5) at 300 ms. 
Before learning, arm trajectory is initialized to minimize the jerk. 

The reward function with the $D$[DOF] arm is as follows:
\begin{eqnarray}
r_t &=& \frac{\sum^{D}_{i=1}\left(D+1-i\right)\left(0.1 f_{i,t}^2+0.5\theta_i^\mathrm{T}\theta_i\right)}{\sum^{D}_{i=1}\left(D+1-i\right)}\\
\Delta r_{\rm 300ms}&=&1.0\times 10^8\left(\left(0.5-x_{\rm 300ms}\right)^2+\left(0.5-y_{\rm 300ms}\right)^2\right)
\end{eqnarray}
, where $x_t$ and $y_t$ are the end-effector position.
DMP \cite{Ijspeert2003Learning} is also used to parameterize the policy.
The parameter settings are as follows: 1000 updates, 10 rollouts, and 100 basis functions.

We summarize the results in \figref{Arm} and \figref{Arm50}. From \figref{ArmCost} and \figref{Arm50Cost}, we can confirm that G-AMDS learns faster than all the other algorithms. 
Moreover, the variance of G-AMDS is smallest.
% タスクも達成できていることがわかる．
As \figref{ArmTraj} and \figref{Arm50Traj} show, it is clear that the G-AMDS agent accomplished both of 10 DOF task and 50 DOF task.
Thus, G-AMDS would have scalability for dimensionality.

\tabref{ArmConv} and \tabref{Arm50Conv} show the average cost and the standard deviation of the cost at the last update.

\begin{figure}[htb]
\centering
  \begin{minipage}[b]{.49\linewidth}
    \centering
    \includegraphics[clip,width=1\linewidth]{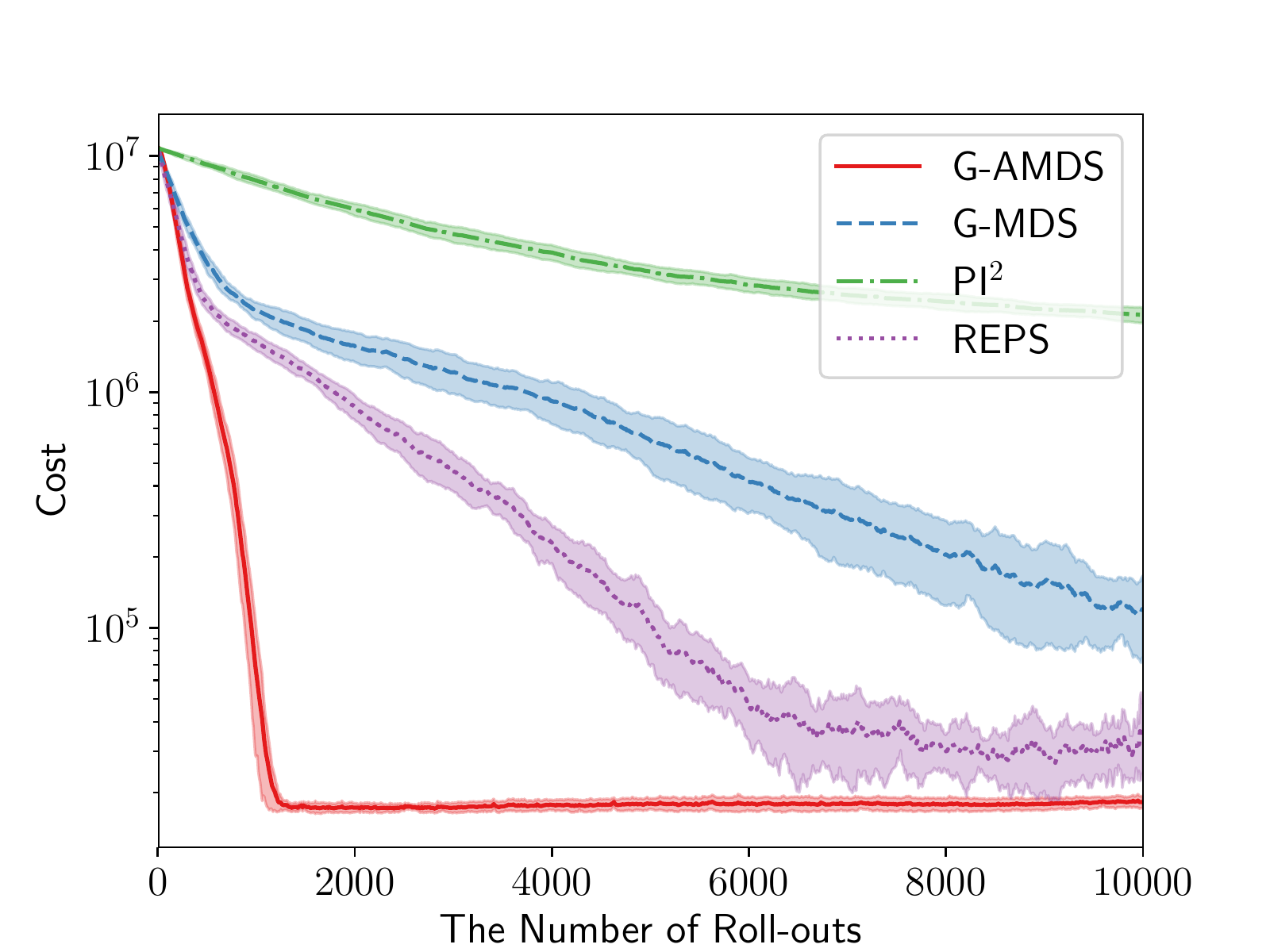}
    \subcaption{Cost}\figlabel{ArmCost}
  \end{minipage}
  \begin{minipage}[b]{.49\linewidth}
    \centering
    \includegraphics[clip,width=1\linewidth]{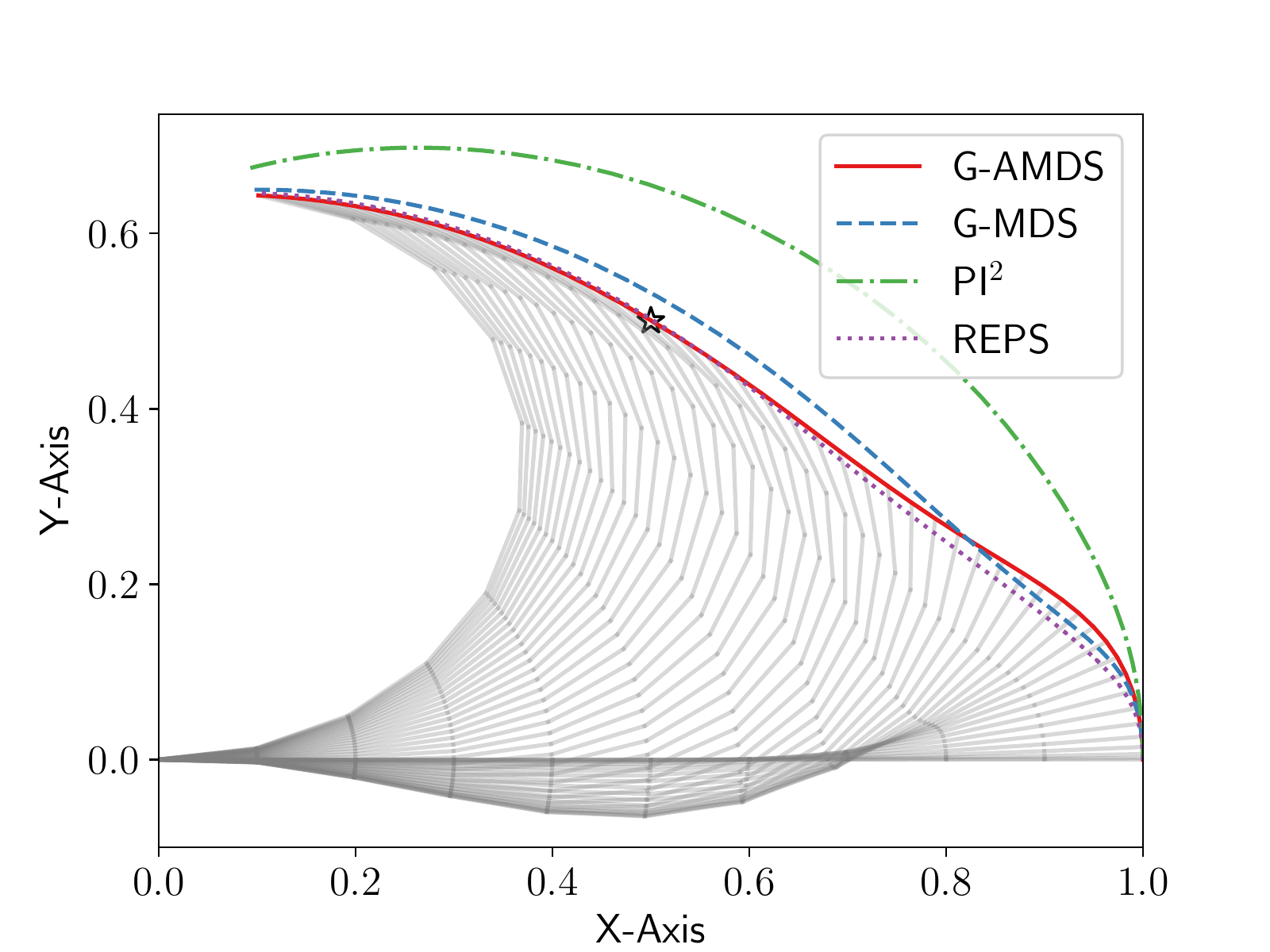}
    \subcaption{Trajectory}\figlabel{ArmTraj}
  \end{minipage}
  \caption{10DOF arm via-point task}\figlabel{Arm}
\end{figure}

\begingroup
\renewcommand{\arraystretch}{2}
\newcolumntype{C}{>{\centering\arraybackslash}X}
\begin{table}[htb]
  \caption{Final cost of 10DOF arm via-point task}\tablabel{ArmConv}
  \begin{tabularx}{\linewidth}{l|C}
    G-AMDS & $1.8\times10^4 \pm 1.1\times10^3$\\
    G-MDS & $1.2\times10^5 \pm 4.6\times10^4$\\
    PI${}^\text{2}$ & $2.1\times10^6 \pm 1.6\times10^5$\\
    Episode-based REPS  & $3.6\times10^4 \pm 1.2\times10^4$\\
  \end{tabularx}
\end{table}
\endgroup

\begin{figure}[htb]
\centering
  \begin{minipage}[b]{.49\linewidth}
    \centering
    \includegraphics[clip,width=1\linewidth]{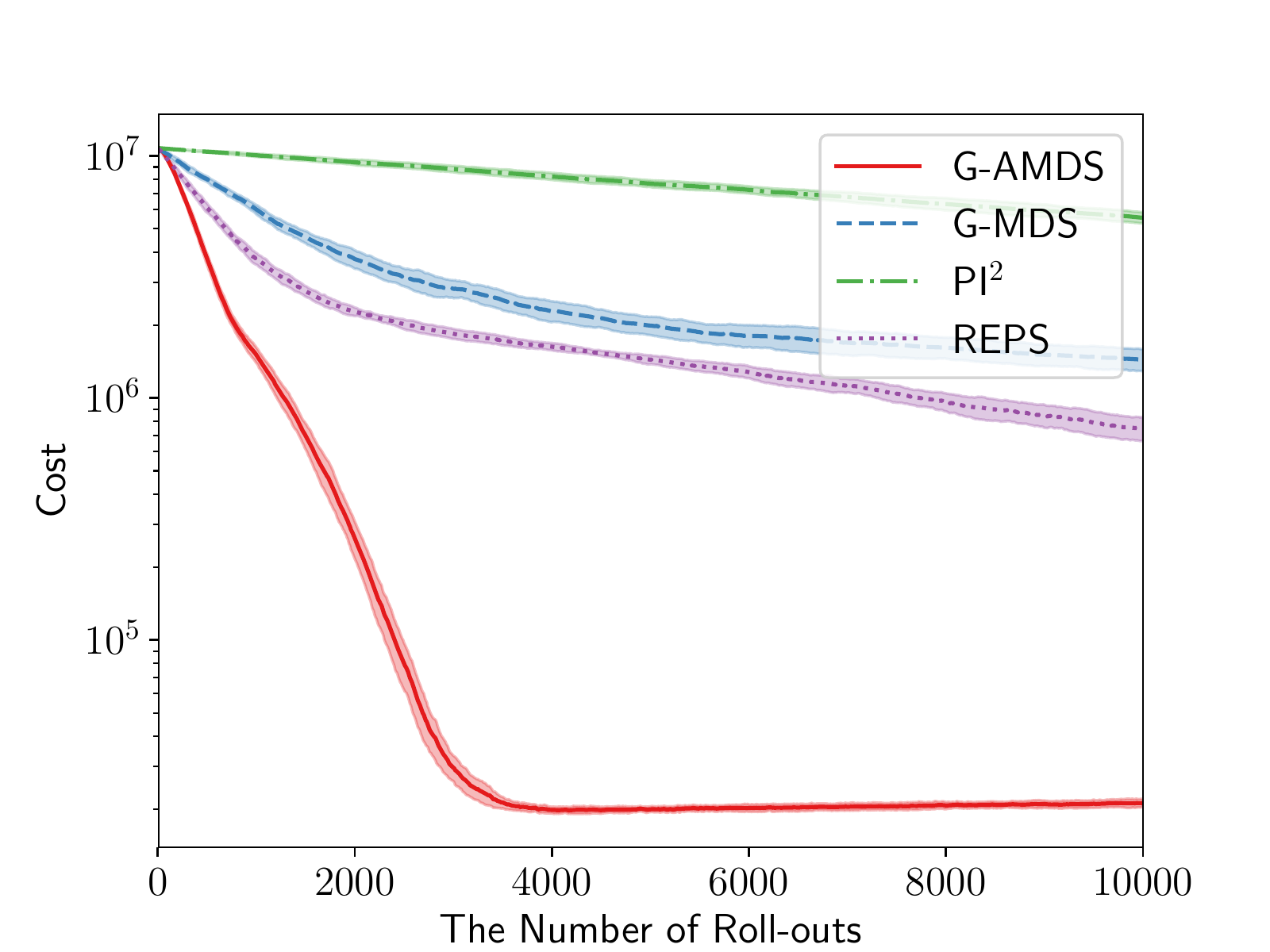}
    \subcaption{Cost}\figlabel{Arm50Cost}
  \end{minipage}
  \begin{minipage}[b]{.49\linewidth}
    \centering
    \includegraphics[clip,width=1\linewidth]{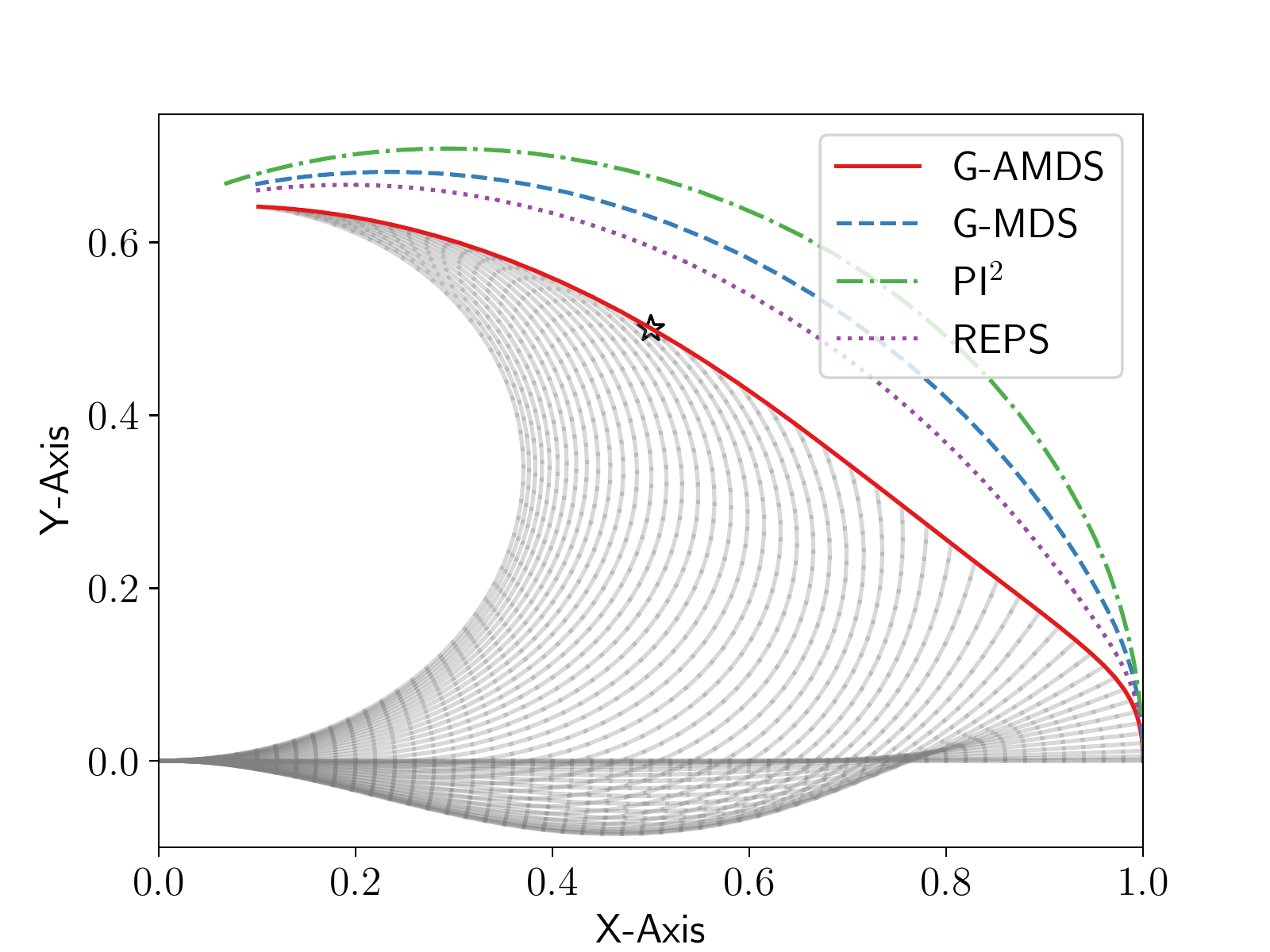}
    \subcaption{Trajectory}\figlabel{Arm50Traj}
  \end{minipage}
  \caption{50DOF arm via-point task}\figlabel{Arm50}
\end{figure}

\begingroup
\renewcommand{\arraystretch}{2}
\newcolumntype{C}{>{\centering\arraybackslash}X}
\begin{table}[htb]
  \caption{Final cost of 50DOF arm via-point task}\tablabel{Arm50Conv}
  \begin{tabularx}{\linewidth}{l|C}
    G-AMDS & $2.1\times10^4 \pm 8.5\times10^2$\\
    G-MDS & $1.4\times10^6 \pm 1.5\times10^5$\\
    PI${}^\text{2}$ & $5.6\times10^6 \pm 3.0\times10^5$\\
    Episode-based REPS  & $7.5\times10^5 \pm 8.3\times10^4$\\
  \end{tabularx}
\end{table}
\endgroup

\section{Relation between MDS and PI${}^\text{2}$}
\label{sec:MDSPI}
% 前章をquantitative evaluations、本章をqualitative evaluationsと言っても良いか?
Theodorou et al. proposed PI${}^\text{2}$ algorithm in \cite{Theodorou2010Reinforcement} and they discussed the relation between PI${}^\text{2}$ and KL control in \cite{Theodorou2012Relative,Theodorou2013information}.
In this section, we provide an explanation of PI${}^\text{2}$ from a viewpoint of MDS.

\subsection{Problem Statement and Algorithm of PI${}^\text{2}$}
\label{subsec:ProbPI2}
We begin with the problem statement of PI${}^\text{2}$ \cite{Theodorou2012Relative, Kappen2011Optimal}:
\begin{eqnarray}
\eqlabel{PI2prob1}
    && \min_{\{\bm u_k\}_{k=t,\dots,T}}\ \mathbf{E}_{\bm \tau}[L(\bm\tau)]\\
\eqlabel{PI2prob2}
    && \mathrm{s.t.}\ \ d{\bm x}_{t}=\bm f(\bm x_t)dt+ \bm G(\bm x_t)(\bm u_t dt+ d\bm w_t), 
\end{eqnarray}
where $\bm x_t\in\mathbb{R}^n$, $\bm f(\bm x_t)\in \mathbb{R}^n$, $\bm G(\bm x_t)\in\mathbb{R}^{n\times m}$, $\bm u_t\in \mathbb{R}^m$ and $\bm \tau:=(\bm x_t, \bm x_{t+dt}, \dots, \bm x_T)$.
$\bm w_t\in\mathbb{R}^m$ is Wiener process \cite{Kappen2011Optimal}. 
It is essential to point out that $\bm u_tdt+d\bm w_t$ plays a role as feedback gain of $\bm G(\bm x_t)$, so our objective is to find the optimal feedback gain; we try to optimize the averaged continuous time series $\bm u_t$ especially.

\eqref{PI2prob2} can be interpreted in two ways. 
Under the model-free reinforcement learning problem, \eqref{PI2prob2} represents the actual physical dynamics of the real plant. 
Under the model-predictive optimal control problem, \eqref{PI2prob2} would represent the predictive model of the real plant. 
% If we have a model of the cumulative cost function $R(\tau)$ additionally, we can solve the problem. 
Essentially, we consider the model-free reinforcement learning setting, below. 

\eqref{PI2prob1}-\eqref{PI2prob2} satisfies linearized Hamilton-Jacobi-Bellman equation under the quadratic cost assumption $L(\bm \tau)=l(\bm\tau)+\sum_{k=t}^T \bm u_k^T\mathbf{R}\bm u_k$ \cite{Theodorou2010Reinforcement, Kappen2011Optimal}. $l(\bm \tau)$ denotes some state dependent cost $l(\bm \tau):=\phi(\bm x_T)+\sum_{k=t}^{T-1} q(\bm x_k)$.
With the path integral calculation, we acquire the analytic solution of the HJB equation.
They finally proposed \algoref{PI2} as an iterative algorithm for the problem. 

Next, we discuss the relation between MDS and PI${}^\text{2}$. 
Theodorou et al. \cite{Theodorou2012Relative, Theodorou2013information} proposed more general problem setting, so we touch the subject in section \ref{subsec:MOREOMD}.

\begin{algorithm}[bth]
\caption{PI${}^\text{2}$ algorithm (see \cite{Theodorou2010Reinforcement} for details)}\algolabel{PI2}      
\begin{algorithmic}[1]
\Initialize{
    parameter vector: $\bm\theta_0:=\bm\theta_\text{init}$\\
    % semi-positive definite matrix: $\bm R$\\
    % immediate cost function: $r_t = q_t+\bm\theta^{\rm T}\bm R\bm\theta$\\
    % terminal cost: $\phi$\\
    % stochastic policy parameter: $\bm g_t^{\rm T}(\bm\theta+\bm\epsilon_t)$\\
    % basis function: $\bm g_t$\\
    % variance $\Sigma_\epsilon$ of mean-zero noise: $\bm\epsilon_t$.
}
\For{$k=1$ to $K$}
    \For{$i=1$ to $m$}
        \For{$t=0$ to $T-1$}
            \State Generate rollout from $\bm\theta_{k-1}+\bm\epsilon_{t,i}$
            \State Compute the projection matrix $\bm M_{t, k}=\frac{\bm R^{-1}\bm g_{t, k}\bm g^{\rm T}_{t, k}}{\bm g^{\rm T}_{t, k}\bm R^{-1}\bm g_{t, k}}$
            \State Evaluate $S\left(\bm\tau_{t, i}\right)=\phi+\sum^{T-1}_{j=t}q_{j, k}+\frac{1}{2}\left(\bm\theta_{k-1}+\bm M_{j, i}\bm\epsilon_{j,i}\right)^{\rm T}\bm R\left(\bm\theta_{k-1} +\bm M_{j,i}\bm\epsilon_{j, i}\right)$
            \State Compute the probability
            % $P\left(\bm\tau_{t, i}\right)=\frac{e^{-\frac{1}{\lambda}S\left(\bm\tau_{t,i}\right)}}{\sum^m_{j=1} \left\{ e^{-\frac{1}{\lambda}S\left(\bm\tau_{t,j}\right)} \right\} }$
            $P\left(\bm\tau_{t, i}\right)=\exp{\bigl(-\frac{1}{\lambda}S\left(\bm\tau_{t,i}\right)\bigr)}/Z$
            \State Compute time dependent differential parameter $\delta\bm\theta_{t}=\sum^m_{i=1}\left[ P\left(\bm\tau_{t,i}\right)\bm M_{t, i}\bm\epsilon_{t,i}\right]$
        \EndFor
        \State Compute time independent differential parameter $\delta\bm\theta=\frac{\sum^{T-1}_{j=0}\left(T-j\right) \delta\bm\theta_j}{\sum^{T-1}_{j=0}\left(T-j\right)}$
    \EndFor
    \State $\bm\theta_{k} = \bm\theta_{k-1} +\delta\bm\theta$
\EndFor
\end{algorithmic}
\end{algorithm}

\subsection{PI${}^\text{2}$ from a Viewpoint of MDS}
\label{subsec:ProbPI22}
% To our best knowledge, 
First of all, we reformulate \eqref{PI2prob1}-\eqref{PI2prob2} as \eqref{PI2MDSprob1}:
\begin{eqnarray}
\eqlabel{PI2MDSprob1}
    && \min_{p(\bm h)}\ \int J(\bm {h}) p(\bm h)d \bm h,%\\
% \eqlabel{PI2MDSprob2}
    % && \mathrm{s.t.}\ \ \{\bm u_k dt+ d\bm w_k\}_{k=1,\dots,T}\sim p(\bm h).
\end{eqnarray}
where $p(\bm h)$ is the probability distribution of the stochastic process $\bm h:=(d\bm z_t, \dots, d\bm z_T)$ with 
%$\bm h:=\{\bm u_kdt+d\bm w_k\}_{k=t,\dots,T}$. 
% $\{\bm h_k\}_{k=t,\dots,T}$ such that $\bm h_t dt = \bm u_tdt+d\bm w_t$. 
$d\bm z_t = \bm u_tdt+d\bm w_t$.
Stochastic process $\bm h$ is the Gaussian process with mean function $\bm\mu:=(\bm u_tdt,\dots, \bm u_Tdt)$ because every increments $d\bm w_t$ are Gaussian.
Once $\bm h_j\sim p({\bm h})$ is sampled, trajectory $\bm \tau(\bm h_j)$ and
$L(\bm\tau(\bm h_j))$ 
are uniquely determined, so we defined $J(\bm h):=L(\bm\tau(\bm h))$. 
% \eqref{PI2prob2} becomes a deterministic difference equation corresponding to it. 
Our problem is to find the optimal probability distribution $p^*(\bm h)$. 

% $J(\bm h)$ such that $\mathbf{E}_{\bm \tau}[R(\bm\tau)]=\int J(\bm {h}) p(\bm h)d \bm h$. 

We introduce MDS approach to optimize $p(\bm h)$: 
\begin{eqnarray}
\eqlabel{PI2MDSprob2}
    p_{k+1}(\bm h)=\argmin_{p(\bm h)}\ \Bigl\{\int J(\bm {h}) p(\bm h)d \bm h+\eta\mathrm{KL}[p(\bm h)\mid p_k(\bm h)]\Bigr\}.
\end{eqnarray}
\eqref{exponentiatedgrad_MDS} is the solution of \eqref{PI2MDSprob2}, which is also known as exponentiated gradient \cite{Shalev-Shwartz2012Online}.
\begin{eqnarray}
\eqlabel{exponentiatedgrad_MDS}
    p_{k+1}(\bm h)=\frac{\exp(-\frac{1}{\eta} J(\bm {h}))p_k(\bm h)}{\int \exp(-\frac{1}{\eta} J(\bm {h}))p_k(\bm h)d\bm h}
\end{eqnarray}
The posterior mean function becomes
\begin{eqnarray}
\eqlabel{expoupdate}
    \label{eq:PI2MDS2}
    \bm\mu_{k+1}&=&\int\bm h\cdot p_{k+1}(\bm h)d\bm h \\
    \label{eq:PI2MDS3}
    &=&\frac{\int\bm h\exp(-\frac{1}{\eta} J(\bm {h}))p_k(\bm h)d\bm h}{\int \exp(-\frac{1}{\eta} J(\bm {h}))p_k(\bm h)d\bm h}.
\end{eqnarray}
By the Monte Carlo approximation, \eqref{PI2MDS3} can be approximated by
\begin{equation}
\eqlabel{Gauss2}
% \tilde{\bm\mu}_k=\frac{\frac{1}{m}\sum^m_{i=1}\bm\theta_i\exp\left(-\eta J_{k-1,i}\right)}{\frac{1}{m}\sum^m_{j=1}\exp\left(-\eta J_{k-1,j}\right)}
\tilde{\bm\mu}_{k+1}=\bm\mu_{k}+\frac{\sum_{j=1}^m (\bm h_j-\bm \mu_k)\exp(-\frac{1}{\eta} J(\bm {h_j}))}{\sum_{j=1}^m \exp(-\frac{1}{\eta} J(\bm {h_j}))}
\end{equation}
where $\bm h_j-\bm \mu_k=(d\bm w_{t,j},\dots,d\bm w_{T,j})$. 
%
%Then we iteratively search the next posterior mean $\bm\mu_{k+2}$ by exploring the neighborhood of $\tilde{\bm\mu}_{k+1}$. 
% The algorithm here can be written such that it resembles that in \algoref{GMDSPI2}. 
With the above mentioned procedure, $p(\bm h)$ gradually gets closer to the optimal $p^*(\bm h)$. 

% Furthermore, from \eqref{Gauss1}, $\bm\theta_i$ for $\bm\epsilon_{k,i}\sim\mathcal{N}\left(0, \bm\Sigma_{\epsilon_k}\right)$  during the update step $k$ can be expressed as follows:
% \begin{equation}
% \bm\theta_i=\bm\mu_{k-1}+\bm\epsilon_{k-1,i}
% \end{equation}
% Substituting this into \eqref{Gauss2}, we have the following:
% \begin{eqnarray}
% \eqlabel{EGPI2}
% \tilde{\bm\mu}_k&=&\!\frac{\sum^m_{i=1}\left(\tilde{\bm\mu}_{k-1}+\bm\epsilon_{k-1,i}\right)\exp\left(-\eta J_{k-1,i}\right)}{\sum^m_{j=1}\exp\left(-\eta J_{k-1,j}\right)}\nonumber \\
% &=&\tilde{\bm\mu}_{k-1}+\sum^m_{i=1}\left(\frac{\exp\left(-\eta J_{k-1,i}\right)\bm\epsilon_{k-1,i}}{\sum^m_{j=1}\exp\left(-\eta J_{k-1,j}\right)}\right).
% \end{eqnarray}

% %% 記号の置き換え
% We here demonstrate that the algorithm is similar to PI${}^\text{2}$. The symbols are replaced as follows:
% \begin{eqnarray}
% P_{k-1,i}&=&\frac{\exp\left(-\eta J_{k-1,i}\right)}{\sum^m_{j=1}\exp\left(-\eta J_{k-1, j}\right)}\\
% M_{k-1,i}&=&1
% \end{eqnarray}

We explain the similarity and difference between \eqref{Gauss2} and PI${}^\text{2}$. 
To simplify the notations, we introduce $\bm\varepsilon_{k,j}:=(\bm h_j-\bm \mu_k)$.
\eqref{Gauss2} becomes
\begin{eqnarray}
\eqlabel{PI2}
\tilde{\bm\mu}_{k+1}&=&{\bm\mu}_{k}
+\sum^m_{j=1}\bm\epsilon_{k,j}P_{k,j}\\
\eqlabel{PI22}
P_{k,j}&:=&\frac{\exp(-\frac{1}{\eta} J(\bm {h_j}))}{\sum_{j=1}^m \exp(-\frac{1}{\eta} J(\bm {h_j}))}
\end{eqnarray}
\eqref{PI2}-\eqref{PI22} correspond to Line 9-13 in \algoref{PI2}. 
There are two important differences between PI${}^\text{2}$ and the algorithms obtained here. 
First, as Line 7-9 in \algoref{PI2} show, PI${}^\text{2}$ sequentially updates the decision variable $\bm\mu_k$ at each time step $t$ based on the provisional cumulative rewards $(S(\bm\tau_{t,1}),\dots, S(\bm\tau_{t,m}))$. 
On the other hand, as \eqref{PI2} shows, our procedure just uses the entire cumulative rewards. 
We can bridge the gap by introducing Dynamic Programming as used in \cite{Theodorou2012Relative,Theodorou2013information} (see \ref{sec:MDSDP}).
% whereas the G-MDS uses only cumulative rewards. 
% Moreover, G-MDS is an algorithm with fewer procedures. 
Second, PI${}^\text{2}$ assumes a Wiener process, but MDS is applicable for arbitrary stochastic processes. 
This difference would be important to deal with more complex stochastic processes. 

\subsection{More General Problem Setting and Online Mirror Descent Trick}
\label{subsec:MOREOMD}
Theodorou et al. proposed more general problem setting in \cite{Theodorou2012Relative,Theodorou2013information}. 
\begin{eqnarray}
\eqlabel{moreprob1}
    \min_{\{\bm u_k\}_{k=t,\dots,T}} && \mathbf{E}_{\bm \tau}[L(\bm\tau)]\\
\eqlabel{moreprob2}
    \mathrm{s.t.} && d{\bm x}_{t}=\bm f(\bm x_t)dt+ \bm G(\bm x_t)\Bigl(d\bm z_t+d\bm\xi_t\Bigr)\\
\eqlabel{moreprob3}
    && d\bm z_t=\bm u_tdt+d\bm w_t. 
\end{eqnarray}
They introduced additional wiener process $d\bm\xi_t$ which represents the stochasticity of passive dynamics. 
All the other variables are defined in section \ref{subsec:ProbPI2} and section \ref{subsec:ProbPI22}.

In this setting, 
trajectory $\bm \tau$ becomes stochastic variable
even after $\bm h_j=(d\bm z_t,\dots, d\bm z_T)$ is sampled.
The evaluated value $J(\bm h)$ is represented by
\begin{equation}
% \bm J(\theta)=\int p\left(h\right)\bm r_\theta \left(h\right)dh.
J(\bm h)=\int p\left(\bm \xi \right) j \left({\bm h}, \bm \xi \right)d\bm \xi,
\eqlabel{expectedJ}
\end{equation}
with $j \left({\bm h}, \bm \xi \right):=L(\bm \tau(\bm h, \bm \xi))$.

It is important to note that we can approximate MDS by: 
\begin{eqnarray}
\eqlabel{OMDSJ}
p_{k+1}(\bm h)&=&\argmin_{ p \in\mathcal{P}}\left\{\int j({\bm h},\bm \xi_k) p(\bm h) d\bm h+\eta B_\phi\left( p(\bm h), p_{k}(\bm h)\right)\right\},\\
\eqlabel{monteJ}
    &&\textrm{s.t.}\ J(\bm h) = \lim_{k\rightarrow\infty}\frac{1}{k}\sum_k j({\bm h},\bm \xi_k)
\end{eqnarray}
% where $\bm j_{k}:=\left[ j_{\bm\theta_1}\left(h_{k}\right),  j_{\bm\theta_2}\left(h_k\right), \cdots\right]$ and $\bigl\{\bm\omega_1, \dots, \bm \omega_k,\dots\bigr\}$
% Once the stochastic variable $\bm \omega_k\sim p(\bm \omega)$ is determined, a candidate evaluation $ j_{\bm h} (\bm \omega_k)$ is uniquely determined as shown in \figref{omd}. 
% It satisfies \eqref{monteJ} with i.i.d. sequence 
We can prove that \eqref{OMDSJ} will asymptotically converge to the optimal solution ~(see \ref{sec:OnlineMirrorDescent}).
This trick is called online mirror descent. 
It enables us to make use of MDS under a single roll-out setting. 
\figref{omd} is the schematic view of \eqref{expectedJ} and \eqref{monteJ}. 
\begin{figure}[htb]
\centering
\includegraphics[clip,width=.5\linewidth]{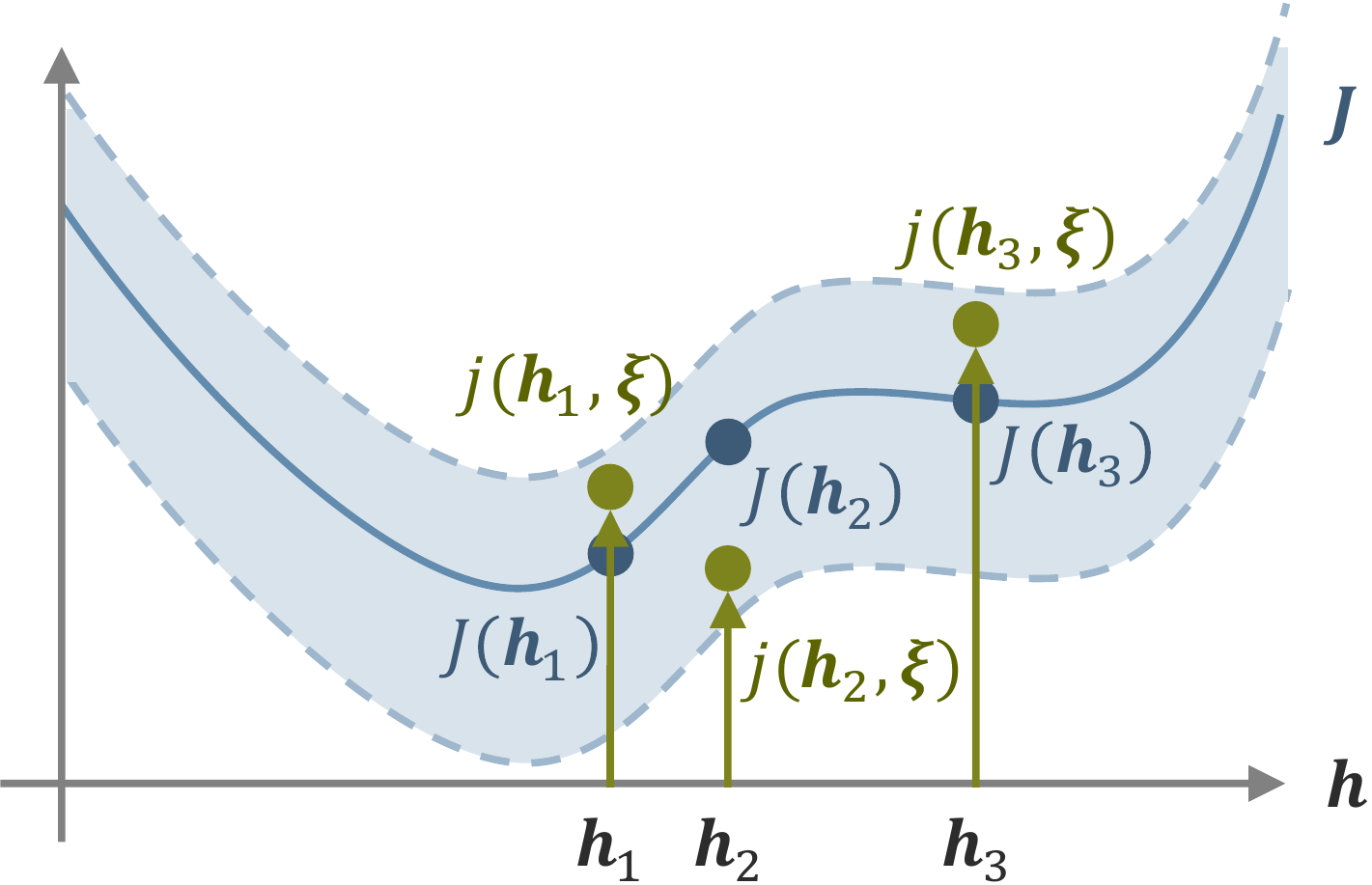}
\caption{The problem setting that evaluation value $J(\bm h)$ is represented as an expectation of $j (\bm h, \bm \xi)$.}
\figlabel{omd}
\end{figure}

Such is the case with reinforcement learning problem. 
We usually employ the expected cumulative reward as the objective function $J(\bm h)$. 
% We define the trajectory as $\bm\tau=\left[\bm x_{t_\text{init.}}, \dots, \bm x_{t_\text{term.}}\right]\in\mathcal{T}$, where $\bm x_t$ represents the state at the time $t$.
% % Assuming dynamic environment from time $t_\text{init.}$ to time $t_\text{term.}$, the trajectory is $\bm\tau=\left[\bm x_{t_\text{init.}}, \dots, \bm x_{t_\text{term.}}\right]\in\mathcal{T}$, where the state is $\bm x_t$ at the time $t$.
% % 強化学習の場合は，このように目的関数が累積コストの期待値となるので，
% % The original objective function $J$ is defined the expected cumulative reward for the generating probability of the trajectory---derived with \eqref{DifObjFunc}---. 
% The objective function becomes
% \begin{equation}
% \eqlabel{DifObjFunc}
% J\left(\bm\theta\right)=\int_{\mathcal{T}^{\bm\theta}} p\left(\bm\tau^{\bm\theta}\right)j\left(\bm\tau^{\bm\theta}\right)d\tau,
% \end{equation}
% where $j\left(\bm\tau^{\bm\theta}\right)$ represents the cumulative reward for the trajectory $\bm\tau^{\bm\theta}\in\mathcal{T}^{\bm\theta}$ under the policy parameter $\bm\theta$.
% Because \eqref{DifObjFunc} behaves like \figref{omd}, 
Although there exist not only uncertainty in the dynamics but also in the reward function, we expect \eqref{OMDSJ} is applicable for the reinforcement learning problems. 

\section{Conclusions}
%% 全体のまとめと結果から言えること
In this research, we proposed four optimization algorithms both for black-box optimization problem and reinforcement learning problem.
On the basis of MD method, we proposed two essential algorithms: MDS and AMDS.
Moreover, we proposed two more approximate algorithms of them: G-MDS and G-AMDS. 
Then, we discussed the relation between our proposed methods and the related algorithms.
Especially in section \ref{sec:MDSPI}, we provided the detailed discussion about the relation between MDS and PI${}^\text{2}$. 
We compared the performances of G-MDS, G-AMDS, PI${}^\text{2}$ and episode-based REPS in two tasks.
G-AMDS showed significant improvements in convergence speed and optimality. 

These results suggest that variety of existing MD extensions can be applied to reinforcement learning algorithms. 
Moreover, it would be also possible that variety of Bayesian techniques such as variational inference are applicable to reinforcement learning algorithms as there exists the theoretical relation between MD method and Bayes theorem~\cite{Dai2016provable}. 
% High-Dimension-Low-Sample-Size(HDLSS) statistics would play a key role in reinforcement learning algorithms. 
We refer to Natural Evolution Strategies (NES)~\cite{Wierstra2014Natural, Salimans2017Evolution}. 
NES uses the natural gradient to update the parameterized distribution. 
% The common points between NES and MDS are an episode-based method and updating the policy distribution.
% On the other hand, they have a different point. 
% NES differentiates an objective function with respect to a parameter of a distribution, while our proposed method differentiates one with respect to a distribution.
% Differentiated by a parameter allows for taking only a parametric distribution, however, Differentiated by a distribution allows for also taking a non-parametric distribution.
% More interestingly, the natural gradient is derived by defining the metric between distributions as KL divergence~\cite{Amari1998Natural}. 
Natural gradient comes from the constraints on KL divergence or Hellinger distance between two distributions.
Because Bregman divergence includes both KL and Hellinger distance, we expect there exists some connections between MDS and NES. 
Recent work suggests that the relation exists between the natural gradient and the MD~\cite{Raskutti2013information}. 
Although we didn't evaluated G-MDS and G-AMDS with the variance-covariance matrix update in this study, we believe CMA-ES and its variants would improve performance. 
Parallelization of MDS algorithms would also be important work.

% Our algorithms would be applied to value-based algorithms if we design the objective function to satisfy Bellman equation. 
% Our algorithms would be wide, as they could have many variations depending on the objective function and the function approximator.
% We consider that using a deep neural network to approximate policy function is more practical, such as TRPO, PPO or NES.

% TODO: 以下のパラグラフはなくても良いかも
% The appropriately choice the learning rate $\eta$ is useful to get the better solution. We should consider this to improve our proposed methods. Furthermore, in \cite{Theodorou2010generalized,Theodorou2010Reinforcement}, the authors use the heuristic normalization $\tilde\eta\frac{J - \min J}{\max J -\min J}$ instead of $\eta J$. We could discuss the heuristics through our proposed methods in future.

% We describe the techniques for performance improvement as future work in the following.

% \textcolor{blue}{同様にKLダイバージェンスを制限する手法であるBayesian Deep Learning / Bayesian Neural Networkを元にすれば，パラメータ数が多い場合も学習可能と言えるかも．}

% DQN etc. Deep Network has a saddle point
% Statistical model \UTF{00B7} Singular model Singlar model (mixed Gauss) has a saddle point

\section*{Acknowledgment}
The research was supported by JSPS KAKENHI (Grant numbers JP26120005, JP16H03219, and JP17K12737).

\appendix
\section{Bregman, KL and RKL divergence}
\label{sec:alphadiv}
We sketch the proof that both of KL and RKL divergence are Bregman divergence~\cite{Amari2009alpha}. 
% `

First of all, we define the smooth convex function $\phi(\bm x)$ in the Bregman divergence \eqref{BregmanDivergence} as 
\begin{equation}
    \phi_\alpha (\bm x)=\frac{2}{1+\alpha}\sum_{i=1}^N\left(1+\frac{1-\alpha}{2}x_i\right)^{\frac{2}{1-\alpha}},
    \label{eq:phialpha}
\end{equation}
where $\bm x\in\mathbb{R}^N$, $x_i>0$ and $\alpha\neq \pm 1$. 
By directly substituting \eqref{phialpha} into the Bregman divergence, we acquire $B_\alpha$. 
The work \cite{Amari2009alpha} provides the proof that $B_\alpha$ becomes $\alpha$-divergence.
The divergence under $\alpha=\pm 1$ condition is defined by a limit case $\alpha\rightarrow\pm 1$. 

The limit case $\alpha\rightarrow\pm 1$ of $B_\alpha$ is easy to calculate. We acquire
\begin{equation}
    \lim_{\alpha\rightarrow +1} B_\alpha(\bm x, \bm y)=\sum_i^N\Bigl[\exp(x_i)-\exp(y_i)+\exp(y_i)\left(y_i-x_i\right)\Bigr],
    \label{eq:forwardalpha}
\end{equation}
and
\begin{equation}
    \lim_{\alpha\rightarrow -1} B_\alpha(\bm x, \bm y)=\sum_i^N \Bigl[-x_i + y_i + (1 + x_i) \log(1 + x_i) - (1 + x_i) \log(1 + y_i)\Bigr].
    \label{eq:reversealpha}
\end{equation}

Under the conditions $x_i=\log p_i$ and $y_i=\log q_i$, \eqref{forwardalpha} becomes
\begin{equation}
    \lim_{\alpha\rightarrow 1} B_\alpha(\log \bm p, \log\bm q)=\sum_i^N q_i\log \frac{q_i}{p_i},
    \label{eq:forwardKLalpha}
\end{equation}
and, under the conditions $x_i=p_i-1$ and $y_i= q_i-1$, \eqref{reversealpha} becomes
\begin{equation}
    \lim_{\alpha\rightarrow -1} B_\alpha(\bm p -1, \bm q - 1)=\sum_i^N p_i\log \frac{p_i}{q_i}.
    \label{eq:reverseKLalpha}
\end{equation}
Here, we used $\sum p_i=\sum q_i=1$ in these calculation.

Finally, we proved both of KL and RKL divergences belongs to Bregman divergence as is shown by \eqref{forwardKLalpha} and \eqref{reverseKLalpha}.

\section{Mirror Descent}
\label{sec:MirrorDescent}
% ここではMirror Descentの説明をする．
% 決定変数$x \in\mathcal{X}$，目的関数 $f:\mathcal{X}\rightarrow\mathbb{R}$， Bregman divergence $B_\phi$と定義する．
We explain the mirror descent algorithm in this section.
Let $x \in\mathcal{X}$ and $f:\mathcal{X}\rightarrow\mathbb{R}$ be a decision variable and an objective function.
\begin{equation}
\eqlabel{MirrorDescent}
x_k=\argmin_{x\in\mathcal{X}}\left\{\langle\nabla f\left(x_{k-1}\right), x\rangle+\eta B_\phi\left(x, x_{k-1}\right)\right\}
\end{equation}
, where $B_\phi$ is the Bregman divergence.
% この第1項において目的関数$f\left(\bm\beta\right)$を近似し，第2項において前回の決定変数$\bm\beta_{t-1}$と離れすぎないよう$\bm\beta\in\mathcal{B}$の更新幅を調整している．
The first term linearlizes the objective function $f\left(x\right)$ around $x=x_{k-1}$, and the second term controls the step size of $x\in\mathcal{X}$ by bounding the Bregman divergence between the new decision variable candidate $x$ and old one $x_{t-1}$.

\section{Accelerated Mirror Descent}
\label{sec:AcceleratedMirrorDescent}
% \textcolor{blue}{TODO: Kricheneの引用を書き，詳しくはKricheneの論文を読めと書く}
% Accelerated Mirror Descent(AMD)の説明をする．
% 決定変数$x \in\mathcal{X}$，目的関数 $f:\mathcal{X}\rightarrow\mathbb{R}$， Bregman divergence $B_\phi$，ハイパーパラメータ$r, \gamma, s$と定義する．
We explain the accelerated mirror descent (AMD) algorithm in this section.
This algorithm is proposed in \cite{Krichene2015Accelerated}.
The AMD is an accelerated method that generalizes Nesterov's accelerated gradient descent.
Let $x \in\mathcal{X}$ and $f:\mathcal{X}\rightarrow\mathbb{R}$ be a decision variable and an objective function.
\begin{eqnarray}
\eqlabel{AMDx}
x_k&=&\lambda_{k-1}{\tilde z}^{k-1}+\left(1-\lambda_{k-1}\right){\tilde x}_{k-1}, \text{with } \lambda_{k-1}=\frac{r}{r+(k-1)}\\
\eqlabel{AMDztilde}
\tilde z_k&=&\argmin_{\tilde z\in\mathcal{X}}\left\{\frac{(k-1)s}{r}\langle\nabla f\left(x_k\right), \tilde z\rangle+B_\phi\left(\tilde z, \tilde z_{k-1}\right)\right\}\\
\eqlabel{AMDxtilde}
\tilde x_k&=&\argmin_{\tilde x\in\mathcal{X}}\left\{\gamma s\langle \nabla f\left(x_k\right), \tilde x\rangle+R\left(\tilde x, x_{k}\right)\right\},
\end{eqnarray}
where $B_\phi$ is the Bregman divergence, $r$ and $\gamma$ are hyper parameters, and $s$ is step-size.
In general, $R\left(x,x'\right)=B_\omega\left(x,x'\right)$ represents the Bregman divergence of the arbitrarily smooth convex function $\omega\left(x\right)$. 
For more detail on the algorithm, refer to~\cite{Krichene2015Accelerated}.

%We here explain the parameters for AMD.
%First we describe the whole formula of AMD and (\ref{eq:AMDx}).
AMD consist of two MD equations Eqs.~(\ref{eq:AMDztilde}) and (\ref{eq:AMDxtilde}).
%Looking at the equation of AMD, (\ref{eq:AMDztilde}) can be regarded as MD. 
Parameter $\lambda$ in \eqref{AMDx} defines the mixture ratio of Eqs.~(\ref{eq:AMDztilde}) and (\ref{eq:AMDxtilde}). 
$\lambda$ is initially close to 1, so AMD behaves according to \eqref{AMDztilde}. 
As $\lambda$ comes close to 0, AMD converges \eqref{AMDxtilde}.

We provide two topics related to this method. 
First, AMD naturally includes simulated annealing, while the existing method such as PI${}^\text{2}$ includes it heuristically~\cite{Theodorou2010generalized, Theodorou2010Reinforcement}.
Parameter $\frac{(k-1)s}{r}$ in \eqref{AMDztilde} is a time-varying learning rate;
as the learning step $k$ proceeds, the factor $\nabla f$ becomes increasingly important for the optimization in \eqref{AMDztilde}. 
This is equivalent to a simulated annealing operation.
It would be more clear if you reformulate \eqref{AMDztilde} in exponentiated gradient form. 

Another topic is about an advantage of reverse-KL (RKL) minimization: $\min_q \mathrm{KL}[q||p]$. 
The methods in this paper and original AMD paper both include it. 
The RKL minimization problem shows mode-seeking behavior
when $p$ is the multi-modal distribution~\cite{Bishop2006Pattern}.
According to \eqref{AMDx}, $x_k$ becomes a multi-modal distribution when $\tilde z_k$ and $\tilde x_k$ are on simplex space. 
Such is the case with $R\left(\tilde x, x_{k}\right)$ in \eqref{AMDxtilde}.
As the learning step $k$ proceeds, $R\left(\tilde x, x_{k}\right)$ gradually becomes to lead $\tilde x$ to $x_{k-1}^{\tilde x}$ from $z_{k-1}^{\tilde x}$. 
We guess the mode-seeking behavior is effective for the AMD to convert to the latter MD algorithm \eqref{AMDxtilde}.

\section{Online Mirror Descent}
\label{sec:OnlineMirrorDescent}
% We define the sampled costs as $\bm j_{k}:=\left[ j_{\bm\theta_1}\left(h_{k}\right), j_{\bm\theta_2}\left(h_k\right), \cdots\right]$ and the mean costs as $\bm J := \left[J\left(\bm\theta_1\right), J\left(\bm\theta_2\right), \cdots \right]$.

We begin with the optimization problem:
\begin{eqnarray}
\bm q_k &=& \argmin_{\bm q\in\mathbb{R}^\infty}\left\{\langle\bm j_{k-1}, \bm q\rangle+\eta B_\phi\left(\bm q, \bm q_{k-1}\right)\right\},
\eqlabel{MDR}\\
% \eqlabel{OMDSJ}
\mathrm{s.t.}\ &&\bm J = \lim_{k\rightarrow\infty}\frac{1}{k}\sum_k \bm j_{k}
\end{eqnarray}

From the formula deformation
\begin{eqnarray}
\bm q_k &=& \argmin_{\bm q\in\mathbb{R}^\infty}\left\{\langle\bm j_{k-1}, \bm q\rangle+\eta\left( \phi\left(\bm q\right)-\phi\left(\bm q_{k-1}\right)-\langle\nabla\phi\left(\bm q_{k-1}\right), \bm q-\bm q_{k-1}\rangle \right)\right\}\\
&=& \argmin_{\bm q\in\mathbb{R}^\infty}\left\{\langle\bm j_{k-1}-\eta\nabla\phi\left(\bm q_{k-1}\right), \bm q\rangle+\eta\phi\left(\bm q\right)\right\},
\end{eqnarray}
and a relational expression of the dual space in mirror descent
\begin{eqnarray}
\nabla\phi\left(\bm q_{k-1}\right)=\nabla\phi\left(\bm q_{k-2}\right)-\frac{1}{\eta}\bm j_{k-2}=\cdots=\nabla\phi\left(\bm q_0\right)-\frac{1}{\eta}\sum_{i=0}^{k-2}\bm j_i,
\end{eqnarray}
we can reformulate \eqref{MDR} as follows:
\begin{eqnarray}
\bm q_k &=& \argmin_{\bm q\in\mathbb{R}^\infty}\left[\langle\bm j_{k-1}-\eta\left\{\nabla\phi\left(\bm q_0\right)-\frac{1}{\eta}\sum_{i=0}^{k-2}\bm j_i \right\}, \bm q\rangle+\eta\phi\left(\bm q\right)\right]\\
&=& \argmin_{\bm q\in\mathbb{R}^\infty}\left[\langle\sum_{i=0}^{k-1}\bm j_i-\eta\nabla\phi\left(\bm q_0\right), \bm q\rangle+\eta\phi\left(\bm q\right)\right]\\
&=& \argmin_{\bm q\in\mathbb{R}^\infty}\left[\langle k\cdot\frac{1}{k}\sum_{i=0}^{k-1}\bm j_i-\eta\nabla\phi\left(\bm q_0\right), \bm q\rangle+\eta\phi\left(\bm q\right)\right],\\
&=& \argmin_{\bm q\in\mathbb{R}^\infty}\left[\langle k\hat{\bm J}-\eta\nabla\phi\left(\bm q_0\right), \bm q\rangle+\eta\phi\left(\bm q\right)\right]
\eqlabel{MDsumR}
\end{eqnarray}

Next, we reformulate the original problem in the same way.
\begin{eqnarray}
\bm q_k &=& \argmin_{\bm q\in\mathbb{R}^\infty}\left\{\langle\bm J, \bm q\rangle+\eta B_\phi\left(\bm q, \bm q_{k-1}\right)\right\}.\eqlabel{MDJ}\\
&=& \argmin_{\bm q\in\mathbb{R}^\infty}\left[\langle k\bm J-\eta\nabla\phi\left(\bm q_0\right), \bm q\rangle+\eta\phi\left(\bm q\right)\right].\eqlabel{MDsumJ}
\end{eqnarray}
% $k$が大きくなるほど，つまり更新回数が増えるほど，$\frac{1}{k}\sum_{i=0}^{k-1}\bm r_i$は$\bm J$に近づく．
% したがって，更新回数がある程度大きい場合，\eqref{MDJ}の$\bm J$を$\bm r_{k-1}$に置き換えて近似することができる．
The more the number of updates $k$ increases, the more $\hat{\bm J}$ gets closer to $\bm J$.
Thus, we can replace $\bm J$ in \eqref{MDJ} with $\bm j_{k-1}$, when the number of updates is sufficient.

% \subsection{Example: Exponentiated Gradient Descent}
% The equation corresponding to \eqref{MDsumR} is as follows:
% \begin{eqnarray}
% \bm q_{k, i} &=& \frac{\exp\left(-\eta j_{k-1,i}\right)q_{k-1, i}}{\sum_{l=1}^\infty\exp\left(-\eta j_{k-1,l}\right)q_{k-1,l}}\\
% &\propto& \exp\left(-\eta j_{k-1,i}\right)q_{k-1, i}=\exp\left(-\eta \sum_{l=1}^{k-1}j_{l,i}\right)q_{0, i}
% \end{eqnarray}

% The equation corresponding to \eqref{MDsumJ} is as follows:
% \begin{eqnarray}
% \bm q_{k, i} &=& \frac{\exp\left(-\eta J_i\right)q_{k-1, i}}{\sum_{l=1}^\infty\exp\left(-\eta J_l\right)q_{k-1,l}}\\
% &\propto& \exp\left(-\eta J_i\right)q_{k-1, i}=\exp\left(-\eta kJ_i\right)q_{0, i}
% \end{eqnarray}

\section{{Dynamic Programming on MDS}}
\label{sec:MDSDP}
% 仮定：確率過程が条件付き確率のチェーンでかけること。離散時間。

We begin with the problem setting: 
\begin{eqnarray}
\label{eq:app1}
    p_{k+1}(\bm h)=\argmin_{p(\bm h)}\ \Bigl\{\int J(\bm {h}) p(\bm h)d \bm h+\eta\mathrm{KL}[p(\bm h)\mid p_k(\bm h)]\Bigr\}.
\end{eqnarray}

We assume discrete time dynamics and the specific Markov Chain structure \cite{Theodorou2013information}:
\begin{eqnarray}
    p_{k}(\bm h)=\prod_{t=1}^T p_k(h_t\mid h_{t-1}).
\end{eqnarray}
In addition, we assume the decomposable objective function
\begin{eqnarray}
    J_(\bm h)=\sum_{t=0}^T F(h_t).
\end{eqnarray}

\eqref{app1} becomes
\begin{eqnarray}
    p_{k+1}(\bm h)=\argmin_{p(\bm h)}\ \Bigl\{
    F(h_0) + \int d\bm hp(\bm h)\sum_{t=1}^T \Bigl(F(h_t)+\eta \log\frac{p(h_t\mid h_{t-1})}{p_k(h_t\mid h_{t-1})}\Bigr)
    \Bigr\}.
\end{eqnarray}
By Bellman principle, we get (see \cite{Theodorou2013information} for details)
\begin{eqnarray}
    V_t(h_t)=\min_{p( h_{t+1}\mid h_t)}\ \Bigl\{
    F(h_t) + \eta \mathrm{KL}[p(h_{t+1}| h_{t})| p_k(h_{t+1}| h_{t})] + 
    \int V_{t+1}(h_{t+1})dp(h_{t+1}| h_{t})
    \Bigr\}.
\end{eqnarray}
Thus we get
\begin{eqnarray}
    p_{k+1}(h_{t+1}|h_t)=\frac{\exp(-\frac{1}{\eta}V_{t+1}(h_{t+1}))p_k(h_{t+1}|h_t)}{Z}.
\end{eqnarray}
and finally, 
\begin{eqnarray}
    \exp(-\frac{1}{\eta}V_{t+1}(h_{t+1}))=\mathbf{E}_{p(h_{k+1}|h_k)}[\exp(-\sum_{k=t}^T F(h_k))].
\end{eqnarray}

\bibliographystyle{elsarticle-num} 
\bibliography{bibtex}


%\begin{thebibliography}{00}


%\bibitem{}

%\end{thebibliography}
\end{document}